\documentclass{iopjournal}
\usepackage[numbers,square,sort&compress]{natbib}

\usepackage{amsmath} 
\usepackage{booktabs} 
\usepackage{multirow} 
\usepackage{color}

\begin{document}

\articletype{Paper} 

\title{OmniFish: A Bioinspired, Modular, Reconfigurable Soft Robotic Fish Enabling Multi-Modal Swimming}

\author{Mathias Jensen$^1$\orcid{0009-0000-5078-3929} and Saravana Prashanth Murali Babu$^1$$^*$\orcid{0000-0003-2024-9391}}

\affil{$^1$SDU Soft Robotics, Biorobotics section, The Mærsk McKinney Møller Institute, University of Southern Denmark, Odense 5230, Denmark}

\affil{$^*$Author to whom any correspondence should be addressed.}

\email{spmb@mmmi.sdu.dk}


\begin{abstract}
In this work, we introduce OmniFish, a bioinspired modular soft robotic fish designed to emulate the full spectrum of body and/or caudal fin (BCF) locomotion modes within a single robotic platform. The design combines morphology averaging across representative fish species with a tendon-driven compliant segmented spine whose effective stiffness can be reconfigured through progressive mechanical jamming. By selectively constraining body segments, OmniFish shifts deformation from whole-body undulation to posterior body bending and localized tail oscillation, enabling bioinspired Anguilliform, Subcarangiform, Carangiform, Thunniform, and Ostraciiform swimming using a single tendon-driven actuator. The design was guided by morphometric analysis and evaluated through caudal-fin thrust characterization, numerical body-drag simulations, kinematic analysis, and swimming performance experiments. Kinematic analysis confirmed the intended robot transition from distributed body undulation to localized tail oscillation. Swimming experiments demonstrated distinct mode-dependent performance, with the Ostraciiform configuration achieving the highest straight-line velocity of 8.14 cm s$^{-1}$ and the lowest cost of transport of 2.42 at 1.5 tail beats s$^{-1}$. These results show that proof-of-concept a mechanically simple stiffness-reconfigurable architecture can generate distinct BCF swimming modes and provide a versatile strategy for developing multimodal soft robotic fish.
\end{abstract}

\section{Introduction}
Coastal and marine environment are under increasing pressure from pollution, habitat degradation, and expanding human activity, creating a growing need for robotic systems that can support underwater monitoring, inspection, and ecological assessment~\cite{rajeshkumar2018studies}. Marine Autonomous Systems such as Autonomous Underwater Vehicles (AUVs) and Remote Operated Vehicles (ROVs) are widely used for tasks such as seabed surveys, infrastructure inspection, and aquaculture monitoring~\cite{wynn2014autonomous}. However, many conventional underwater vehicles rely on rigid bodies and propellor- or jet-based propulsion which can generate turbulence, re-suspend sediment, disturb benthic organisms, damage aquatic vegetation, and produce noise that may affect marine life~\cite{Fileman_deMora_Vance_2020, cai2025measuring}. Their rigid architectures and limited maneuverability can constrain operation in shallow, cluttered, or ecologically sensitive coastal environments~\cite{springer2016Handbook}. These limitations motivate the development of alternative underwater propulsion systems that are more complaint, efficient, and maneuverable with low-disturbance. Shaped by natural selection, Fish provide a compelling biological model for such robotic systems. 

\begin{figure}[htbp]
    \centering
    \includegraphics[width=\linewidth]{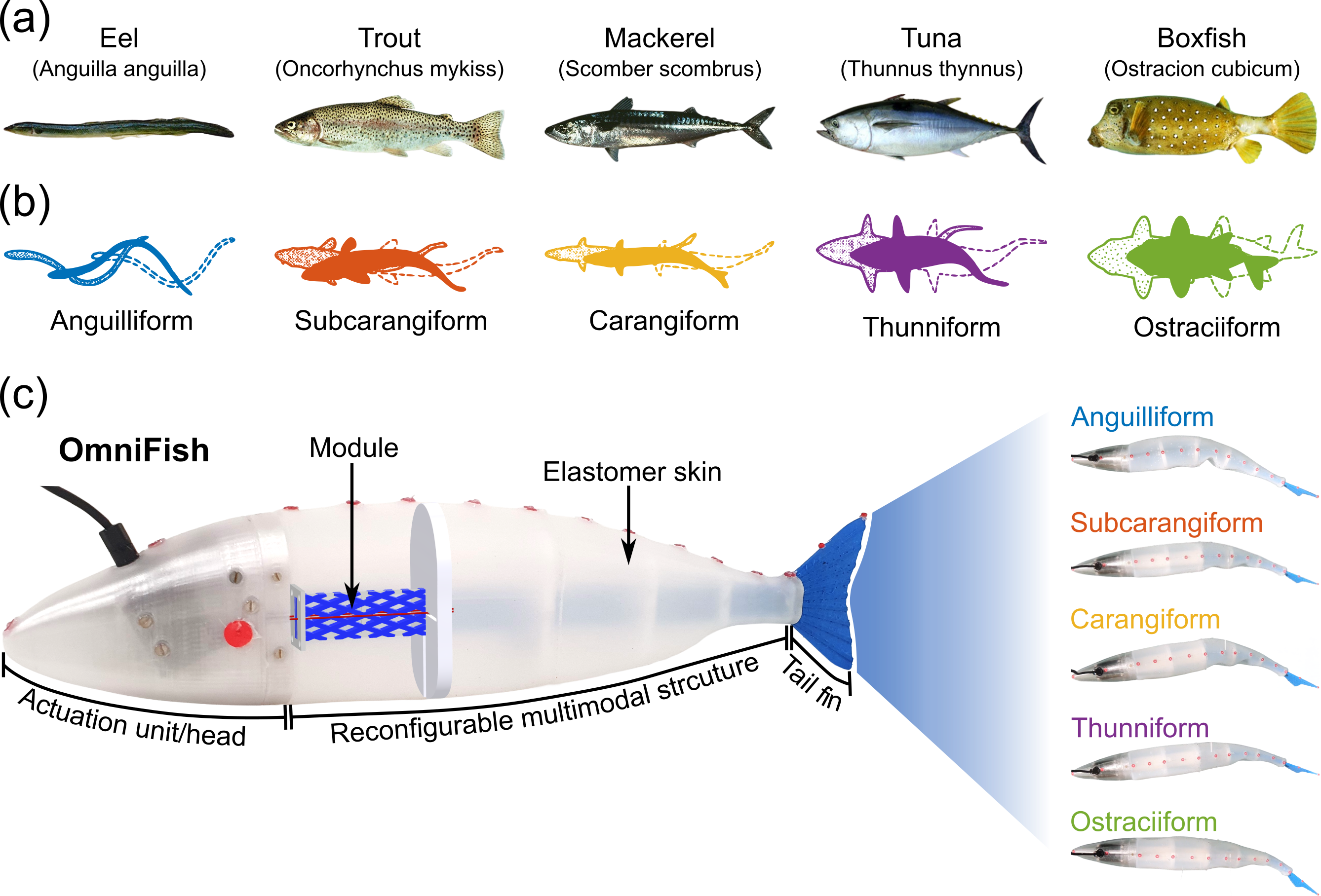}
    \caption{(a) Examples of fish using BCF locomotion. (b) BCF modes aligned with the fish using the correspondent mode. (c) The proposed soft robot OmniFish, capable of reproducing full spectrum BCF locomotion.}
    \label{fig:intro}
\end{figure}

Fish achieve efficient and adaptive underwater locomotion by coordinating flexible spines, distributed muscle actuation, and complaint fins to interact with the surrounding flow~\cite{fish2020advantages,webb1984form} and has long inspired studies of hydrodynamics and swimming mechanics. Most fish generate thrust using body and/or caudal fin (BCF) propulsion, where body waves propogate posteriorly and drive the caudal fin (tail), while other species rely on median/and or paired fin (MPF) locomotion. Propulsion modes are further divided into undulatory motion, with waves propagating along the propulsive structure, or oscillatory motion, where the propulsive structure swivels on its base without wave formation. BCF propulsion is commonly classified into four undulatory modes—Anguilliform, Subcarangiform, Carangiform, and Thunniform—and one oscillatory mode, Ostraciiform (see Figure~\ref{fig:intro}a for different fish using BCF and Figure~\ref{fig:intro}b for illustration of modes). Anguilliform involves undulatory waves traveling along the entire length of the fish’s body. Subcarangiform and Carangiform rely on posterior body undulations. Thunniform primarily utilize high-efficiency caudal-fin oscillation with minimal body bending, whereas Ostraciiform is characterized by pendulum-like oscillation of the caudal fin with a relatively rigid body~\cite{sfakiotakis1999review}. Fish-inspired locomotion represents a broad design space with the potential to provide efficient, maneuverable, quiet operation, and low-wake propulsion, making it attractive for underwater robots intended to operate in sensitive or complex aquatic environments~\cite{triantafyllou2000hydrodynamics}. Its efficiency arises in part from the exploitation of unsteady vortex dynamics, including the recovery of energy from self-generated and neighboring wake structures~\cite{LAUDER2005425}. In addition, fish-like propulsion provides agile maneuvering in cluttered and turbulent habitats~\cite{sfakiotakis1999review}, highlighting its relevance for sustainable marine robot design.

Bio-inspired fish robotics have advanced substantially since the swimming mode taxonomoy of Sfakiotakis, Lane, and Davies established the framework for classifying BCF based aquatic locomotion. Early robotic fish, such as the modular \textit{Amphibot}~\cite{crespi2005amphibot} and Princeton's mechanical robotic lamprey~\cite{hultmark2007flowfield} achieved eel-like undulation using rigid, servo-driven bodies and traveling-wave central pattern generators (CPGs). More recent developments have increasingly shifted from rigid to soft robotic fish, which offer continuous body curvature, structural compliance, and motion patterns closer to biological swimmers to emulate the BCF swimming modes. Soft pneumatic eel robots generate traveling waves via phase-shifted chamber inflation along a flexible backbone achieving Anguilliform motion~\cite{milana2020eelworm,nguyen2022anguilliform}, while electrohydraulic dielectric elastomer actuator (DEA) have produced transparent, eel-larva-like swimmers that undulate silently in salt water~\cite{christianson2018translucent}. Subcarangiform and Carangiform swimming is exemplified by MIT’s hydraulically actuated \textit{SoFi}, which demonstrated untethered operation in coral reefs using fluidic elastomer actuators and an onboard pump~\cite{katzschmann2018exploration}. Alternatives include compact antagonistic shape memory alloy (SMA) springs for quiet but slower tail actuation~\cite{muralidharan2021development}, high-frequency artificial DEA muscles for fast oscillation~\cite{shintake2018soft, wang2023design}, and a Pangasius-inspired design with a servo-driven fin-ray flexible tail~\cite{youssef2022design}. Thunniform locomotion has been realized by the rigid-soft hydrid \textit{Tunabot}, which utilized a rigid head and a soft posterior with ribs, a keeled peduncle, and flexible skin~\cite{zhu2019tuna}. Its successor \textit{Tunabot flex}~\cite{white2021tunabot} highlighted the benefits of stiffness tuning for efficiency, which is also supported by~\cite{zhong2021tunable}. \textit{OpenFish} extends these concepts into an active-passive tail optimized for speed~\cite{van2022openfish}, while \textit{CarbonFish} introduced a bistable carbon fiber-reinforced plastic (CFRP) and silicone mechanism for rapid actuation~\cite{xiong2024carbonfishbistableunderactuated}. Ostraciiform robots, typically boxfish-inspired, remain mostly rigid-bodied with oscillating soft or elastic fins, such as \textit{MARCO}~\cite{kodati2008microautonomous}.

Despite these advances, most robotic fish are primarily designed around a single dominant swimming mode. This specialization can improve performance for a particular mode, but it limits adaptability across different flow conditions, operating spaces, and mission requirements. Distributed undulation can provide greater body compliance and may support maneuvering, interaction with confined environments, and adaptation to disturbed flows, whereas posteriorly localized bending can reduce unnecessary body motion and favor efficient forward propulsion. A robotic fish capable of shifting between these deformation patterns could therefore adapt its locomotion strategy according to operating conditions instead of remaining optimized for a single task. Recent studies have begun to address this by combining multiple BCF locomotion modes within a single robotic platform, including systems based on soft bionic DEA driven muscles~\cite{wang2025soft} and tensegrity structures with vacuum actuated jamming muscles~\cite{wang2026adaptive}. However, these approaches have primarily focused on the four undulatory BCF modes, while the integration of all five BCF modes, including Ostraciiform locomotion, within a single mechanically simple soft robotic fish remains underexplored. A key challenge is therefore to design a robotic architecture that can systematically shift deformation from distributed whole-body undulation to localized tail oscillation without requiring multiple independent actuators or complex actuation schemes.

In this work, we present \textit{OmniFish}, a bioinspired modular soft robotic fish designed to emulate the full spectrum of BCF locomotion modes within a single robotic platform (Figure~\ref{fig:intro}c). The design combines morphological averaging across representative fish species with a tendon-driven compliant segmented spine whose effective stiffness can be reconfigured through progressive mechanical jamming. We hypothesized that by selectively constraining body segments, OmniFish would shift deformation from whole-body undulation to posterior body bending and localized tail oscillation, enabling BCF-inspired Anguilliform, Subcarangiform, Carangiform, Thunniform, and Ostraciiform swimming using a  single tendon-driven actuator and that this would alter the balance between body deformation losses and caudal propulsion, resulting in mode-dependent differences in swimming speed and energetic cost. We evaluated the system through morphometric analysis, caudal-fin thrust characterization, numerical body-drag simulations, kinematic analysis, swimming performance experiments, and cost-of-transport (COT) analysis. The results demonstrate that \textit{OmniFish} can generate distinct BCF-inspired swimming modes within a mechanically simple tendon-driven architecture, with mode-dependent differences in swimming speed and COT, providing a versatile design strategy for multimodal soft robotic fish.

\section{Method}
\subsection{Hydrodynamic Operating Regime}
\label{sec:hydro}
To define the robot morphology and operating parameters, five representative fish species were selected: eel \textit{Anguilla anguilla}, trout \textit{Oncorhynchus mykiss}, mackerel \textit{Scomber scombrus}, tuna \textit{Thunnus thynnus}, and boxfish \textit{Ostracion cubicum}. Together, these species span the major body and/or caudal fin (BCF) locomotion modes considered in this study (Figure~\ref{fig:intro}a-b). Specifically, the eel represents Anguilliform swimming, the trout Subcarangiform, the mackerel Carangiform, the tuna Thunniform, and the boxfish Ostraciiform.
The swimming performance of the selected species spans a broad range of speeds. Eels exhibit an optimal sustained swimming speed of 0.74-1.02~BL~s$^{-1}$~\cite{palstra2008cost}, while trout, mackerel, and tuna sustain average swimming speeds of 2.92, 3.14, and 1.30~BL~s$^{-1}$~\cite{webb1971swimming,sambilay1990interrelationships}. In boxfish, BCF locomotion is primarily associated with burst-and-coast swimming at average speeds of approximately 5.57~BL~s$^{-1}$~\cite{hove2001boxfishes}. These reference values give an operational range of approximately 0.5-3~BL~s$^{-1}$ for sustained biological BCF swimming. This range was used to guide the hydrodynamic design of the robot and selection of experimental operating parameters, such as actuation frequency range, rather than representing an expected performance target for the prototype. For a body length (the standard length, SL, is used as body length for this calculation, see Supplementary information S1 for more information on morphological measurements) of 0.4~m, this corresponds to an absolute swimming speed of 0.2-1.2~m~s$^{-1}$.

Fish swimming is commonly characterized using dimensionless parameters that describe the hydrodynamic regime and oscillatory propulsion. Two particularly relevant parameters assessed here are: the Reynolds number, the Strouhal number. The Reynolds number (equation~\ref{eq:reynolds}) quantifies the ratio of inertial to viscous forces in the surrounding flow, where $U$ is swimming velocity, $l$ a reference length, and $\nu$ the kinematic viscosity of water. Fish typically swim in regimes with Reynolds numbers in the range of approximately $\text{Re}=10^5\text{--}10^7$. The Strouhal number characterizes the relationship between oscillatory tail motion to forward swimming speed, (equation~\ref{eq:stouhal}), where $f$ is the tail-beat frequency and $H$ is the peak-to-peak caudal fin amplitude. In this work tail-beat frequency is represented as tail beats pr second (TB~s~$^{-1}$ equivalent to international standard unit for frequency hertz, Hz). Efficient biological swimmers are commonly associated with the Strouhal numbers in the range of approximately $\text{St}=0.2\text{--}0.4$~\cite{arakeri2009fluid}.
\begin{equation}
    \text{Re} = \frac{Ul}{\nu}
    \label{eq:reynolds}
\end{equation}
\begin{equation}
    \text{St} = \frac{fH}{U}
    \label{eq:stouhal}
\end{equation}
With SL=0.4~m and a swimming velocity of 0.2-1.2~m~s$^{-1}$, the corresponding Reynolds number is $\text{Re} = 0.8 \times 10^5\text{--}4.8 \times 10^5$. This is at the lower end of the biological swimming regime while still within an inertially dominated flow regime. The target peak-to-peak caudal fin amplitude was set to 0.15–0.21~BL, consistent with cross-species biological observations~\cite{di2021convergence}. Combining this amplitude range with an efficient Strouhal range of 0.24–0.35 yields an estimated tail-beat frequency of approximately 0.6–7~TB~s~$^{-1}$ across the operational speeds.

These estimates define the hydrodynamic design envelope for the robotic platform and guided the selection of body length, caudal fin amplitude, and actuation frequency used in the subsequent fin characterization and swimming experiments.
\subsection{Morphometric Analysis}
\label{sec:morph}
After defining the hydrodynamic operating regime, the robot morphology was derived from representative images of fish specimen collected from FishBase (www.fishbase.se). It involved tracing the outline of the fish in the images to get representative contours where morphometrics could be measured and body points exported. The image-processing workflow is described in detail in Supplementary information S1.

Lateral profile images of each species (Figure~\ref{fig:intro}a) were manually annotated. These images were then binarized, and normalized with respect to standard length (SL), enabling direct comparison of relative morphology independent of absolute body size. Outer contours were extracted. All contours were oriented clockwise and subdivided into body and caudal fin contours. Each contour was further separated into dorsal and ventral segments. These segments were resampled to $N = 256$ points by cumulative arc-length parameterization~\cite{walter1996approximate}, as defined in equation~\ref{eq:arc}, ensuring uniform point spacing along the contour. The resampled segments were stitched together, to obtain smooth, evenly sampled body and caudal-fin outlines for each species.
\begin{equation}
    s(t) = \int_0^t \sqrt{\dot{x}(\tau)^2 + \dot{y}(\tau)^2} \; d\tau, \: \: \: 0 \le t \le 1
    \label{eq:arc}
\end{equation}

Direct transverse cross-sectional imaging was unavailable for all specimens except the boxfish, which was processed using the same contour-extraction workflow as the the lateral profiles. Qualitative cross-sectional descriptions were instead obtained from FishBase: the eel is described as circular, the trout as laterally compressed, and the mackerel and tuna as oval. Cross sections were approximated as ellipses centered at the origin using the standard parametric representation of an ellipse in equation~\ref{eq:ellipse} where $a$ and $b$ denote the semi-major and semi-minor axes, respectively. The major axis was aligned with the dorso–ventral ($y$) direction and the minor axis with the lateral ($x$) direction. For the eel, $a=b$, corresponding to a circular cross section. For the trout, the minor axis was set to 40\% of the major axis to represent a compressed cross section, while for the mackerel and tuna it was set to 50\% of the major axis to represent an oval cross section.
\begin{equation}
    (x,y) = \left(a \, \cos{(t)}, b \, \sin{(t)} \right), \: \: \: 0 \le t \le 2\pi
    \label{eq:ellipse}
\end{equation}

An average body outline, caudal fin shape, and cross section was then computed by pointwise averaging of the corresponding resampled contours across the selected species using equation~\ref{eq:average} where $N$ is the number of species included in the morphometric dataset. The pointwise averaging procedure was used to define a common morphology for a platform intended to investigate multiple BCF locomotion modes. It was not intended to produce a mathematically or hydrodynamically optimal body shape. Selecting a single species-specific morphology would make the proportions of that species the fixed geometric reference for all five locomotion modes. The averaging approach instead provided a common geometry derived from all five selected body forms, allowing the effect of stiffness distribution to be examined without choosing one species as the sole morphological baseline.
\begin{equation}
    (x(t),y(t)) = \frac{1}{N} \sum_{i=1}^N \bigl(x_i(t),y_i(t)\bigr)
    \label{eq:average}
\end{equation}
The resulting real world specimen and averaged contours defined the geometric design envelope for the robotic platform and guided the scaling of body and caudal fin morphology used in the subsequent CAD design.
\subsection{Caudal Fin Characterization}
\label{sec:fins}
\begin{figure}[htbp]
    \centering
    \includegraphics[width=\linewidth]{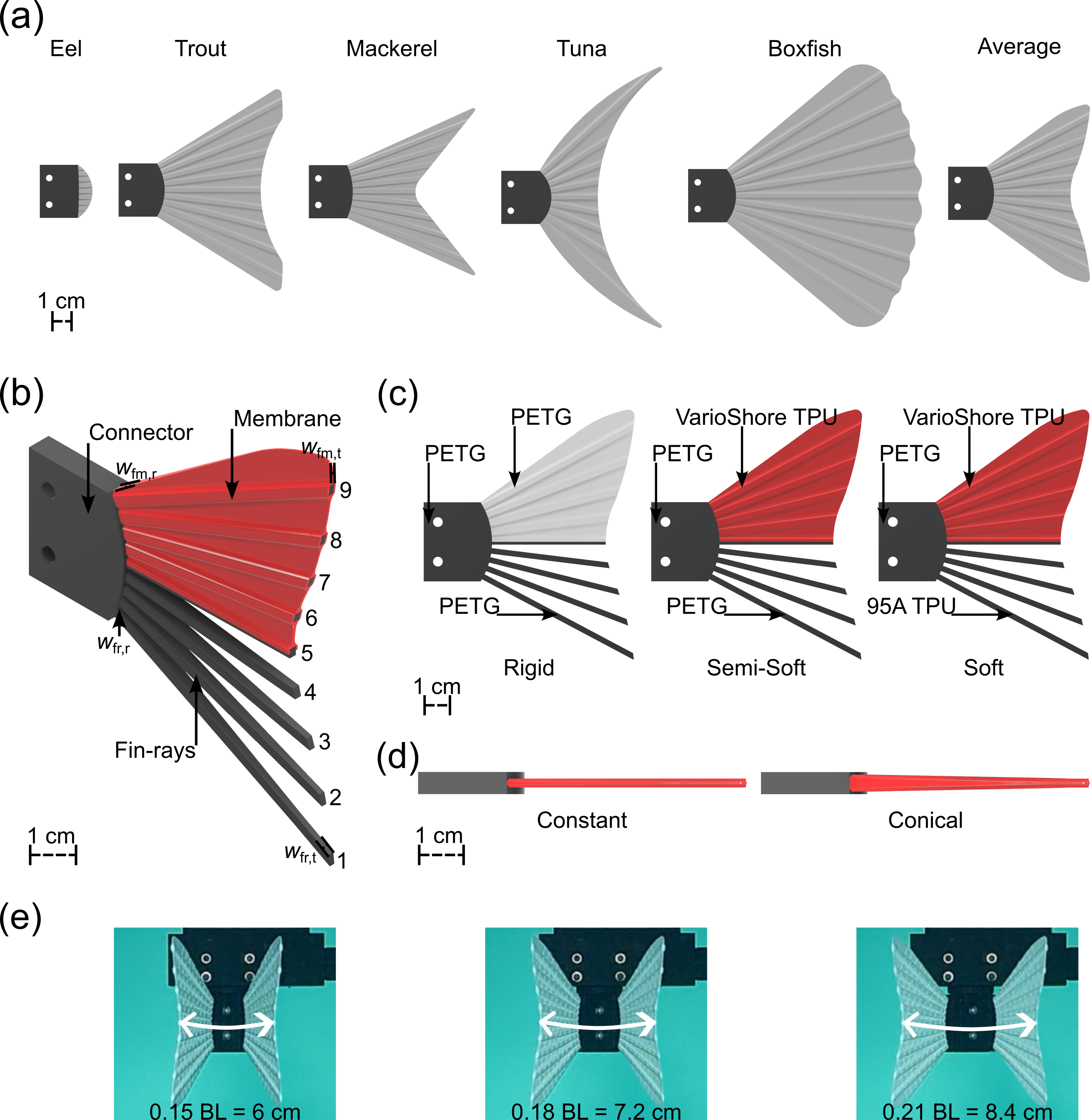}
    \caption{Details on the fin design, dimensional references are available. (a) Six caudal fins CAD designs (b) Structure of the caudal fin design, corresponding of a connector, a caudal fin membrane, and nine fin rays. (c) Three caudal fin material combinations; completely rigid, semi-soft, soft fins. (d) Two thickness profiles; constant and variable conical profiles. (e) Snapshots from experiments showing the three amplitude levels.}
    \label{fig:fins}
\end{figure}

The caudal fin is the primary propulsive structure in most BCF swimmers and plays a central role in thrust generation. Therefore, a systematic caudal fin characterization was performed to identify a fin morphology and structural configuration suitable for integration into the robotic platform. The morphometric profiles extracted in Section~2.2 were uniformly scaled to match the 40~cm robotic platform and six caudal fins were designed in computer aided design (CAD) software (Figure~\ref{fig:fins}). The CAD models represent idealized versions of the extracted fin contours: minor geometric features were smoothed or simplified, and all fins were enforced to be symmetric about the dorso-ventral axis to facilitate part design and manufacturing, while preserving the defining morphological characteristics of each species (Figure~\ref{fig:fins}a). All the fins share common caudal peduncle geometry to enable standardized attachment. As a consequence of this standardization, the mackerel and tuna fins exhibit slightly widened profiles relative to the extracted contours, whereas the trout and boxfish fins were marginally narrowed.

Each caudal fin consists of an internal fin-ray skeleton encapsulated by a continuous fin membrane as indicated on Figure~\ref{fig:fins}b. The skeleton comprises nine fin-rays evenly distributed across the full span of the fin, providing structural support for transmitting propulsion, while the membrane defines the external fin shape and contributes to overall complaince of the structure. To investigate the effects of material stiffness and structural compliance on thrust generation, each fin morphology was fabricated using three distinct material combinations: (i) rigid fin-rays with a rigid fin membrane, (ii) rigid fin-rays with a soft fin membrane, and (iii) soft fin-rays with a soft fin membrane. Each material combination was additionally produced with two thickness-profiles (see Figure~\ref{fig:fins}c-d). The constant-thickness profile employed fin-rays of $w_{\text{fr,r}}=w_{\text{fr,t}}=0.8$~mm thickness and a membrane of $w_{\text{fm,r}}=w_{\text{fm,t}}=0.4$~mm thickness along the entire fin. The variable-thickness (conical) profile tapered from $w_{\text{fr,r}}=2.4$~mm fin-rays and $w_{\text{fm,r}}=1.2$~mm membrane thickness at the fin root to $w_{\text{fr,t}}=0.8$~mm fin-rays and $w_{\text{fm,t}}=0.4$~mm membrane thickness at the distal edge. Compared with the constant-thickness configuration, the conical profile introduces an additional spanwise stiffness gradient along the fin. In total, this design space comprised 6 morphologies $\times$ 3 material combinations $\times$ 2 thickness profiles, resulting in 36 unique caudal fins. All fins were manufactured using a multi-material 3D printing process, more information can be found in Supplementary information S2.

Thrust characterization was performed using a lever-arm-based experimental setup (see Supplementary information S2) inspired by the micro-thrust measurement apparatus proposed by \textit{D. Hu et al.}~\cite{hu2021micro}. The caudal fin was mounted at the lower end of a lever arm and oscillated in water. During actuation, the thrust generated by the fin produced a moment about a central fulcrum. This moment was balanced by an opposing force measured by a load cell mounted on the opposite side of the lever arm. Under static moment equilibrium, the thrust force was calculated from the measured load according to equation~\ref{eq:lever} where $d_{\text{load cell}}$ and $d_{\text{thrust}}$ denote the respective moment arms.
\begin{equation}
    F_{\text{load cell}} d_{\text{load cell}} = F_{\text{thrust}} d_{\text{thrust}}
    \label{eq:lever}
\end{equation}

Each fin was evaluated across 14 actuation frequencies (0.6, 1.0, 1.5, 2.0, 2.5, 3.0, 3.5, 4.0, 4.5, 5.0, 5.5, 6.0, 6.5, and 7.0 Hz) and 3 trailing-edge amplitudes (0.15, 0.18, and 0.21 BL, see Figure~\ref{fig:fins}e) covering the frequency and amplitude ranges defined in Section~\ref{sec:hydro}. For each fin–frequency–amplitude condition, measurements were repeated six times (n = 6). In total, the dataset comprised 36$\times$14$\times$3$\times$6 = 9,072 thrust measurements, these can be seen in summarized Supplementary information S3.
\subsection{Numerical Body-Drag Simulation}
\label{sec:sim}
A streamlined body morphology is essential for reducing hydrodynamic drag and improving swimming performance. To compare the hydrodynamic characteristics of the investigated fish-inspired body shapes, numerical body-drag simulations were performed using Computational Fluid Dynamics (CFD). The simulations were used as a comparative design tool to evaluate relative differences in drag force, pressure loading, wake development, and off-axis force generation across the candidate body geometries. 

CFD simulations were performed with the open-source finite volume software OpenFOAM. The simulations were conducted under steady-state, incompressible flow conditions using the simpleFoam solver, which is commonly used for external flow problems~\cite{versteeg1995finitevolume}. Turbulence effects were modeled using the ($k\text{-}\omega$) shear stress transport (SST) model, which provides improved performance for flows involving adverse pressure gradients and separation~\cite{menter1994two}.

\begin{figure}[htbp]
    \centering
    \includegraphics[width=\linewidth]{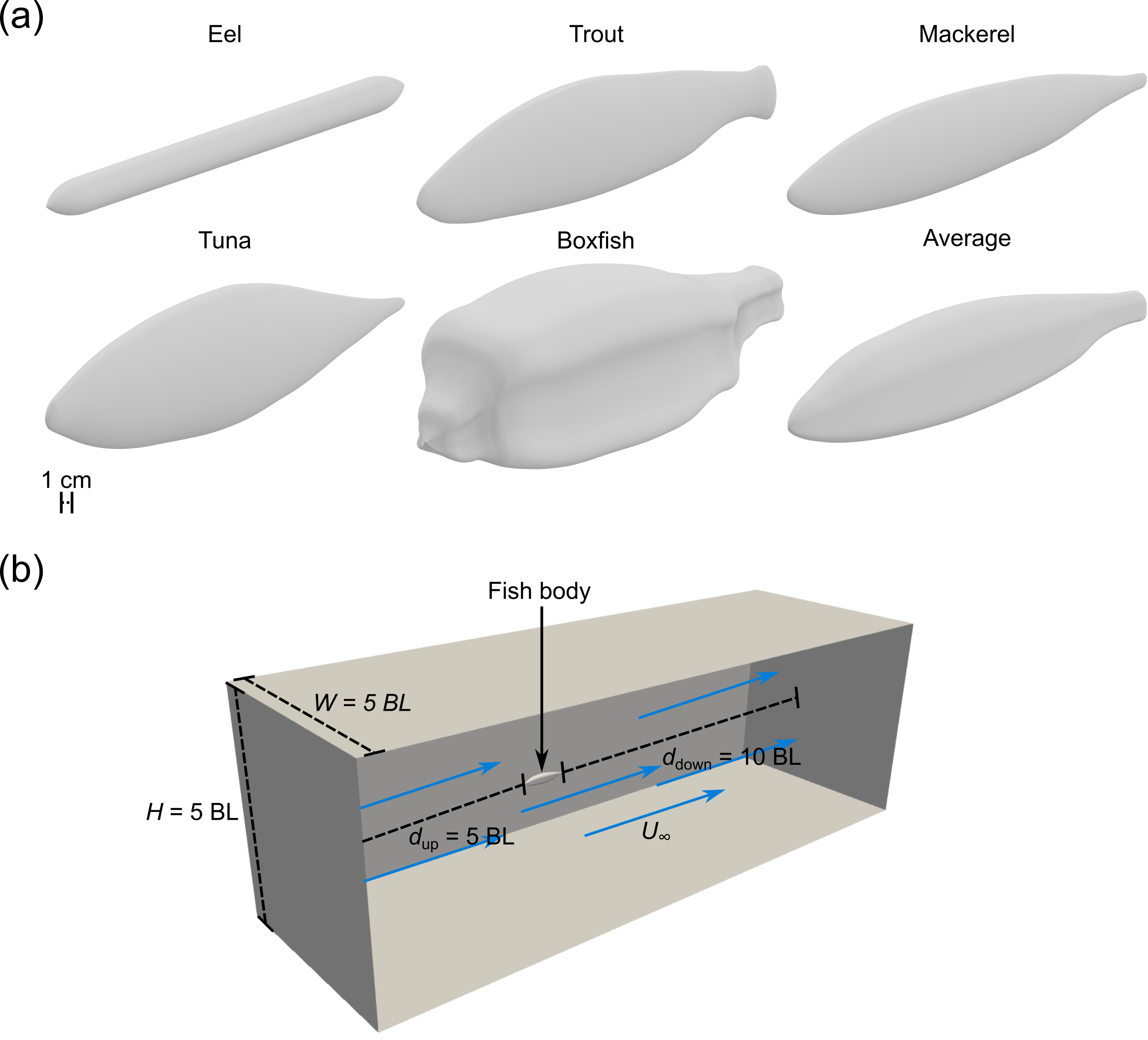}
    \caption{Details on body drag simulation. (a) Six fish bodies CAD design, with reference length. (b) Computational domain used in OpenFOAM simulations showing inlet flow direction $U_\infty$, fish body location, and upstream/downstream and lateral distances relative to BL.}
    \label{fig:drag-sim}
\end{figure}

Six body geometries were designed in CAD (see Figure~\ref{fig:drag-sim}a) using the morphometric body and cross sectional profiles extracted in section~\ref{sec:morph} and uniformly scaled to match the 40 cm robotic platform as in section~\ref{sec:fins}. $N=128$ point body contours were connected with splines, smoothing the geometries, but keeping the defining morphological characteristics of each species, and then extruded. $N=32$ point cross sectional contours were also connected with splines and distributed throughout the length of the body and a loft was used to carve out the final 3 dimensional shape. The geometries were exported as triangulated surface meshes (STL format). Each geometry was placed inside a rectangular computational domain representing an external flow environment (Figure~\ref{fig:drag-sim}b). The dimensions of the domain were chosen sufficiently large to minimize boundary effects, following standard external flow guidelines: upstream distance: 5 BL, downstream distance: 10 BL, lateral clearance: 5 BL. A uniform inlet velocity ($U_\infty$) was prescribed, while a zero-gradient condition was applied at the outlet. The lateral, top, and bottom boundaries were treated as symmetry planes, and the fish body surface was modeled as a no-slip wall. Water was used as the working fluid with constant density ($\rho = 1000$ kg m$^{-3}$) and kinematic viscosity ($\nu = 1\times10^{-6}$ m$^2$ s$^{-1}$).

The computational mesh was generated in two steps: 1. A structured background mesh was created using \textit{blockMesh}. 2. The mesh was refined and conformed to the geometry using \textit{snappyHexMesh}, which generates body-fitted meshes from triangulated surfaces. Surface refinement levels were systematically varied to perform a mesh independence study. Five refinement levels were considered ((3,4), (4,5), (5,6), (6,7), (7,8)), corresponding to increasing mesh density near the body surface. The resulting mesh sizes ranged from approximately (O($10^5$)) to (O($10^6$)) cells. Mesh quality was assessed using the \textit{checkMesh} utility, ensuring acceptable values of non-orthogonality, skewness, and cell aspect ratio. The mesh independence study compared the converged drag force across successive mesh refinements, and the mesh was considered sufficiently resolved when changes in drag were below approximately 1–3\%. Refinement level $(5,6)$ was selected as the optimal compromise between numerical accuracy and computational cost, and therefore was used for all subsequent simulations. More information of the mesh independence study can be found in Supplementary information S4.

A uniform inlet velocity was imposed, and multiple flow speeds (0.2, 0.4, 0.6, 0.8, 1.0, 1.2~m~s$^{-1}$) were investigated to assess the dependence of drag on Reynolds number corresponding to operating range defined in Section~\ref{sec:hydro}. Simulations were run until residuals of velocity and turbulence quantities iteratively decreased below predefined thresholds and the drag force reached a steady value. Flow fields were post-processed using ParaView. The wake structure was analyzed using, velocity magnitude contours on mid-plane slices and pressure distribution on the body surface for the 1.2~m~s$^{-1}$ case. The pressure coefficient was computed from the OpenFOAM kinematic pressure field using equation~\ref{eq:presnorm} where \(p_\infty\) is the reference pressure and \(U\) is the free-stream velocity. Here, \(p_\infty\) was taken from the far-field pressure in the \(U=1.2~\mathrm{m\,s^{-1}}\) case. These visualizations provide insight into flow separation and wake development, which are directly related to pressure drag and overall hydrodynamic performance~\cite{Anderson2011Fundamentals}.

\begin{equation}
    C_p = \frac{p - p_\infty}{0.5\,U^2},
    \label{eq:presnorm}
\end{equation}

From the simulation the drag force acting upon the body was extracted together with its pressure and viscous force components. The drag coefficient was computed, using the projected frontal area of each body, with equation~\ref{eq:drag-coeff} where $F_D$ is the drag force, $\rho$ the fluid density, $U$ the flow velocity, and $A$ the frontal area of the body.

\begin{equation}
    C_D = \frac{2F_D}{\rho U^2 A}
    \label{eq:drag-coeff}
\end{equation}

Since \textit{OmniFish} is intended to emulate the full range of BCF swimming modes, an additional numerical analysis was performed to assess the sensitivity of each body to off-axis flow. To approximate the local body misalignment that occurs during undulatory swimming, each body was simulated at incidence angles of \(5^\circ\) and \(10^\circ\) at a flow rate of 0.6~m~s$^{-1}$. This flow rate was chosen since it is the middle of the velocity range, and the maximum velocity every mode might experience. The resulting drag increases and side forces were then extracted from the simulation.
\subsection{Kinematic Analysis}
\label{sec:kin}
After selecting the caudal fin and body morphologies \textit{OmniFish} was build, then kinematic analysis was performed to evaluate whether \textit{OmniFish} with progressive segment jamming could produce systematic changes in body deformation corresponding to the five intended BCF-inspired swimming modes. The analysis consisted of two steps: first, quantifying the relationship between commanded servo angle and resulting caudal fin peak-to-peak amplitude; and second, quantifying full-body midline kinematics in terms of amplitude, phase, and curvature.

Fifteen red markers were distributed along the midline of \textit{OmniFish}, with marker numbering starting at the snout and increasing posteriorly (see Supplementary information S5). The robot was submerged in a water tank while being attached to the linear rig and constrained in the forward direction. A camera was positioned vertically above the robot to record the top view, where all midline markers remained visible throughout swimming. Additional illumination was provided using a projector light source. The experimental setup is shown in Supplementary Information S5. Experiments were performed for 15 swimming cycles at tail-beat frequencies of 0.6, 1.0, 1.5, and 2.0~TB~s$^{-1}$ with servo amplitudes ranging from 10--78$^\circ$ (see Supplementary information S6). Video recordings were acquired at a resolution of 1080p and a frame rate of 60~fps. The communication cable was lightly guided during experiments to maintain the robot within the camera field of view while minimizing disturbance to the motion.

Initial kinematic analysis focused on the relationship between servo amplitude and resulting caudal fin peak-to-peak displacement. The marker located at the caudal tip was selected for tracking because it represented the point of maximum lateral excursion. The caudal tip marker was tracked using Tracker (Tracker 6.2.0, Douglas Brown, Wolfgang Christian, Robert M Hanson) over the first six swimming cycles. Within the software, the coordinate system was aligned with the longitudinal axis of the robot with its origin positioned at the snout. A calibration stick corresponding to the full body length of the robot was used to convert pixel coordinates to physical dimensions. The first swimming cycle was excluded due to startup transients. Peak-to-peak amplitudes were subsequently extracted. A complete investigation of amplitude response for all swimming modes is provided in Supplementary Information S6. These results guided the selection of servo actuation amplitudes and operating frequencies used throughout the swimming experiments.

After selecting the control parameters, full-body midline kinematics were analyzed for each mode actuated at 1~TB~s$^{-1}$. All 15 markers were tracked for five representative locomotion cycles. Continuous body midlines were generated by first applying cumulative arc-length parameterization, as defined in equation~\ref{eq:arc}. Small lateral translations and rotations were introduced by the experimental constraint and communication cable. To obtain a consistent coordinate system, the instantaneous orientation of each midline was defined by the vector connecting the snout and caudal-fin tip. Each midline was rotated so that this vector was aligned with the longitudinal $x$-axis. The mean midline motion was then calculated across the five cycles. Fifteen equally spaced temporal phases were selected from the cycle-averaged motion for visualization of the progressive body deformation. At every normalized body position, amplitude was calculated independently for each cycle from the maximum excursion on lateral $y$-axis. The mean amplitude envelope and standard deviation were calculated across the five complete cycles. Head and tail amplitude were defined as amplitude from 0.00--0.22~BL (head) and 0.88--1~BL (tail), respectively. The location of maximal amplitude and the ratio of head:tail amplitude were also quantified. Following the Fourier-based approach of Akanyeti and Liao~\cite{akanyeti2013kinematic}, lateral motion at each body position was represented by its fundamental harmonic, from which the local phase was extracted. Phase was subsequently unwrapped along the body to evaluate posterior propagation of the deformation wave. Curvature along the body was quantified as the reciprocal of the local radius of curvature, following the approach of Di Santo \textit{et al.}~\cite{di2021convergence}. Curvature was evaluated from the spatial derivatives of the smoothed midline coordinates with equation~\ref{eq:curvature} and expressed in BL$^{-1}$. Mean absolute curvature was used to quantify the local bending magnitude.

\begin{equation}
\kappa=
\frac{x_s y_{ss}-y_s x_{ss}}
{\left(x_s^2+y_s^2\right)^{3/2}}
\label{eq:curvature}
\end{equation}
\subsection{Swimming Performance Characterization}
\label{sec:swim}

Swimming experiments were performed to quantify the locomotion performance and energetic cost of \textit{OmniFish} across the five BCF swimming modes. The robot was attached to a low-friction linear carriage suspended above the tank using dual aluminum profiles, allowing longitudinal translation while constraining lateral motion (see Supplementary information S5). A marker was placed on the side of the rigid head and recorded with a camera capturing the swimming trajectory.

To get a measure of efficiency, the voltage supplied to the servo and the voltage across a 0.15~$\Omega$ shunt resistor were measured. Servo current was determined from the measured shunt voltage according to equation~\ref{eq:current} where $V_{\text{shunt}}$ is the measured voltage across the shunt resistor and $R_{\text{shunt}}$ its resistance. The electrical power consumption during swimming was subsequently calculated using equation~\ref{eq:power}.

\begin{equation}
I_{\text{servo}} =
\frac{V_{\text{shunt}}}{R_{\text{shunt}}}
\label{eq:current}
\end{equation}

\begin{equation}
P =
V_{\text{servo}}I_{\text{servo}}
\label{eq:power}
\end{equation}

Swimming efficiency was quantified using the dimensionless Cost of Transport (COT), calculated using equation~\ref{eq:cot} where $P$ is mean electrical power consumption, $m$ the robot mass, $g$ gravitational acceleration, and $U$ the mean swimming velocity.

\begin{equation}
COT=
\frac{P}{mgU}
\label{eq:cot}
\end{equation}

The processed swimming data were additionally used to characterize the operating hydrodynamic regime through the Reynolds number and Strouhal number enabling comparison of the robot operating regime with biological swimmers.

Swimming experiments were performed for all five modes over a travel distance of approximately 50~cm at tail-beat frequencies of 0.6, 1.0, 1.5, and 2.0~TB~s$^{-1}$ with six repetitions per condition, resulting in a total of $5\times4\times6=120$ trials. The communication cable was guided vertically upwards with sufficient slack to minimize interaction forces. The head marker was tracked throughout the full video duration. A tape measure positioned alongside the tank was used for spatial calibration. Startup and deceleration regions corresponding to the first and last 5~cm of travel were excluded from analysis. Consequently, only the central region spanning 5--45~cm was considered (see Supplementary information S7). Swimming velocity was obtained from the temporal derivative of the tracked position using local linear regression over a moving window spanning three tail-beat cycles. This window length was selected to reduce short-term fluctuations while preserving changes in swimming behaviour. A nominal swimming velocity was estimated as the 75th percentile of the local velocity distribution. Steady swimming was subsequently identified by retaining data points exceeding 35\% of this nominal velocity. Additional morphological filtering was applied to remove short disturbed bursts and to fill brief interruptions in otherwise steady swimming intervals. This procedure excluded occasional disturbances caused by cable interaction or increased carriage friction. Electrical power measurements were synchronized with the kinematic measurements through temporal alignment and smoothing prior to analysis. Mean swimming velocity and mean electrical power consumption were subsequently calculated over valid swimming intervals only. The processed data were subsequently used to quantify swimming speed and COT across locomotion modes and actuation frequencies, and to compare both within-mode and between-mode performance.

\section{Results}
\subsection{Morphometrics of Six Fish Bodies}

\begin{figure}[!htbp]
    \centering
    \includegraphics[width=\linewidth]{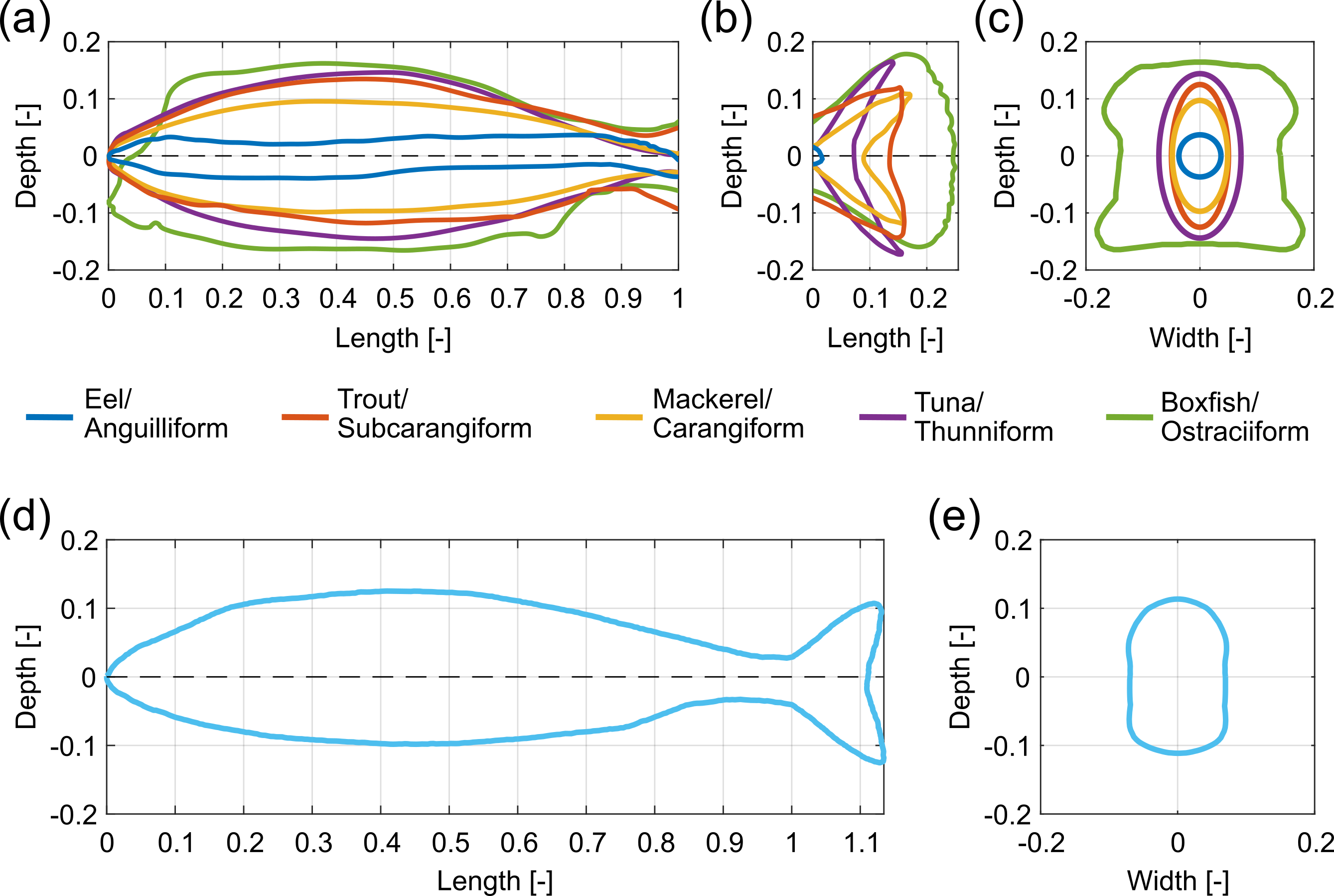}
    \caption{Morphometric contours extracted from representative fish species spanning the BCF locomotion spectrum. (a)-(c) Body outlines, caudal fin contours, and cross-sectional profiles derived from the five selected species. (d)-(e) Averaged body, caudal fin, and cross-sectional geometries obtained by pointwise averaging of the normalized contours.}
    \label{fig:fish-model}
\end{figure}

Using the contour-extraction and normalization procedure described in Section~\ref{sec:morph} body, caudal fin, and cross-sectional profiles were obtained for the representative fish species spanning the BCF locomotion spectrum. The resulting contours are shown in Figure~\ref{fig:fish-model}a-c and reveal clear morphology-dependent differences associated with each locomotion mode.

The eel exhibits an elongated and slender body with minimal body depth, no distinct caudal fin, and a circular cross section, consistent with Anguilliform swimming. In contrast, the boxfish displays a deep, rigid body with a broad caudal peduncle (see Supplementary information S1 for more information on fish morphology terms), a large caudal fin, and a wide box-like cross section, reflecting the morphology associated with Ostraciiform locomotion. The Subcarangiform, Carangiform, and Thunniform swimmers (trout, mackerel, tuna) display progressively more streamlined profiles, with the tuna exhibiting the narrowest peduncle and shortest, widest caudal fin, typical of high-performance cruising swimmers.

\begin{table}[!htbp]
    \centering
    \caption{Morphometrics extracted from length-normalized silhouettes of representative fish for each BCF swimming mode. All quantities are expressed relative to SL where applicable. “n.a.” indicates that a distinct peduncle is not present.}
    \begin{tabular}{l c c c c c c}
        \hline
         &  Eel & Trout & Mackerel & Tuna & Boxfish & Average \\
         \hline
         \textbf{Body metrics} \\
         \hline
         Total length, TL & 1.01 & 1.16 & 1.17 & 1.16 & 1.25 & 1.14 \\
         Body depth, BD & 0.062 & 0.250 & 0.192 & 0.288 & 0.326 & 0.223 \\
         Body width, BW & 0.062 & 0.100 & 0.097 & 0.144 & 0.361 & 0.143 \\
         Frontal area, A$_{\text{F}}$ & 0.0042 & 0.0195 & 0.0147 & 0.0323 & 0.0963 & 0.0281 \\
         Fineness ration, F & 16.08 & 4 & 5.18 & 3.47 & 3.07 & 4.48 \\
         Peduncle length, PL & n.a. & 0.157 & 0.031 & 0.019 & 0.150 & 0.080\\
         Peduncle depth, PD & n.a. & 0.095 & 0.033 & 0.029 & 0.100 & 0.041 \\
         Peduncle slenderness, PS & n.a. & 0.381 & 0.173 & 0.099 & 0.307 & 0.184 \\
         \hline
         \textbf{Caudal metrics} \\
         \hline
         Caudal span, CD & n.a. & 0.264 & 0.219 & 0.338 & 0.338 & 0.232\\
         Caudal length, CL & n.a. & 0.160 & 0.164 & 0.155 & 0.250 & 0.135 \\
         Caudal area, A$_{\text{C}}$ & n.a. & 0.0302 & 0.0143 & 0.0148 & 0.0608 & 0.0194 \\
         Caudal aspect ration, AR & n.a. & 2.31 & 3.36 & 7.75 & 1.88 & 2.78 \\
         \hline
    \end{tabular}
    \label{tab:morphometric}
\end{table}

Point-wise averaging of the five species-specific contours produced the average body and caudal fin geometry shown in Figure~\ref{fig:fish-model}d. The averaged body  retained a streamlined profile with an elongated caudal peduncle and a slightly forked, relatively short caudal fin of similar length to that of the trout. The averaged cross section (Figure~\ref{fig:fish-model}e) is moderately wide and slightly box-like, while retaining overall elliptical characteristics. Overall it retains many of the efficient swimmers streamlined bodies, and smaller forked caudal fin but still exhibits some of the boxfish characteristics mostly seen in its cross section. Table~\ref{tab:morphometric} summarizes the critical morphometrics extracted from the six bodies and caudal fins. These metrics guided later CAD design. Supplementary information S1 show how different metrics were extracted.
\subsection{Caudal Fin Thrust Characterization}
\label{sec:fin-results}
\begin{figure}[!htbp]
    \centering
    \includegraphics[width=\linewidth]{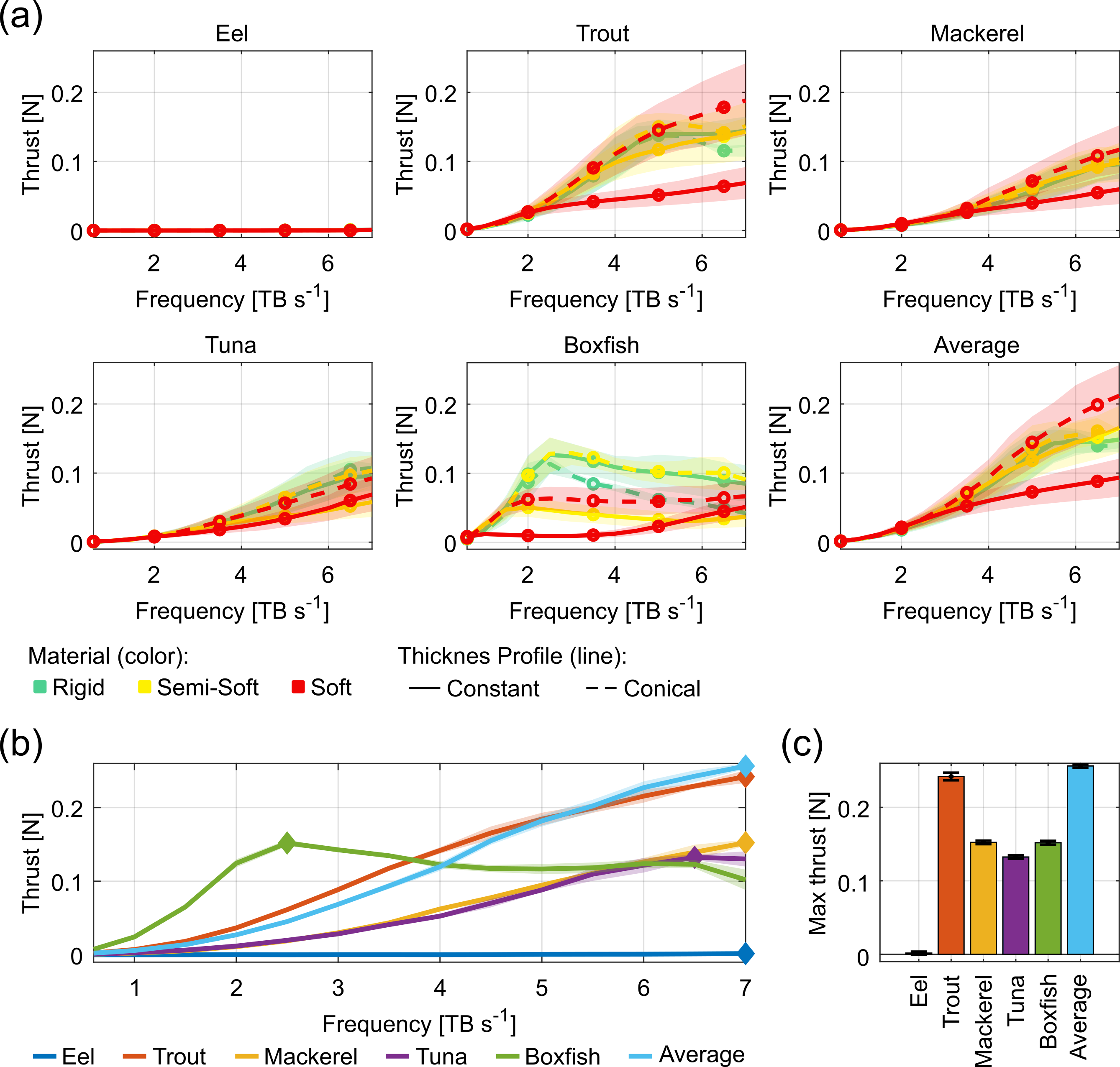}
    \caption{Caudal fin thrust characterization across morphology, material composition, thickness profile, actuation frequency, and amplitude. 
(a) Frequency-dependent thrust response for each fin type. Line color denotes material combination, line style indicates thickness profile, and shaded regions represent the range across amplitudes. 
(b) Thrust curves for the optimal configuration of each fin type; diamond markers indicate maximum thrust. 
(c) Maximum thrust achieved at the optimal operating condition for each fin.}
    \label{fig:fin-thrust-results}
\end{figure}

Based on the morphometric analysis, six caudal fin geometries were designed and fabricated as described in Section~\ref{sec:fins}. Figure~\ref{fig:fin-thrust-results} summarizes the thrust generated across actuation frequencies for all fin morphologies, material combinations, thickness profiles, and amplitude levels. The full thrust signals are provided in Supplementary Information~S3, and representative high-performing fin responses are shown in Supplementary Video~S1. Figure~\ref{fig:fin-thrust-results}a shows the frequency response for each fin type, where individual curves corresponds to material-thickness combinations, and shaded regions indicate the range of thrust values across the tested amplitudes. To quantify the effects of the experimental parameters such as fin morphology, material composition, thickness profile, actuation frequency, and amplitude on thrust generation, a linear mixed-effects (LME) model (equation~\ref{eq:lme}) was employed. The mixed-effects ANOVA results are summarized in Table~\ref{tab:lme_anova}. Fin type was the only significant main effect, indicating that caudal fin morphology played a dominant role in determining thrust generation. In contrast, frequency, amplitude, material, and thickness did not show significant main effects when considered independently. These results indicate that the effect of actuation frequency, amplitude, material composition, and thickness profile depends strongly on fin morphology rather than following a single common trend across all fins as shown in Figure~\ref{fig:fin-thrust-results}a.
\begin{equation}
    \text{Thrust $\sim$ FinType $\times$ Material $\times$ Thickness $\times$ Frequency $\times$ Amplitude + (1$|$FinID)}
    \label{eq:lme}
\end{equation}

\begin{table}[htbp]
\centering
\caption{Mixed-effects ANOVA results for thrust. Only statistically relevant and physically interpretable terms are shown.}
\label{tab:lme_anova}
\begin{tabular}{lcccc}
\hline
Term & F-statistic & DF$_1$ & DF$_2$ & p-value \\
\hline
FinType & 6.77 & 5 & 7560 & $2.63 \times 10^{-6}$ \\

Frequency & 0.26 & 13 & 7560 & 0.996 \\
Amplitude & 0.001 & 2 & 7560 & 0.999 \\

Material & 0.018 & 2 & 7560 & 0.983 \\
Thickness & 0.0002 & 1 & 7560 & 0.988 \\

FinType $\times$ Frequency & 2833.7 & 65 & 7560 & $< 0.001$ \\
FinType $\times$ Amplitude & 2.15 & 10 & 7560 & 0.018 \\

Material $\times$ Frequency & 0.075 & 26 & 7560 & 1.000 \\
Thickness $\times$ Frequency & 0.056 & 13 & 7560 & 1.000 \\

FinType $\times$ Material $\times$ Frequency & 690.6 & 130 & 7560 & $< 0.001$ \\
FinType $\times$ Thickness $\times$ Frequency & 40.93 & 65 & 7560 & $< 0.001$ \\

FinType $\times$ Frequency $\times$ Amplitude & 151.9 & 130 & 7560 & $< 0.001$ \\

\hline
\end{tabular}
\end{table}

Across all fin types, thrust generally increases with actuation frequency; however, the nature of this increase varies significantly. Some fins (e.g., trout, tuna, and average) exhibit a near-monotonic increase, while others (notably the boxfish) show an early peak followed by a plateau or reduced performance at higher frequencies. This diversity in response is consistent with the absence of a significant main effect of frequency in the LME model, and the presence of strong FinType $\times$ Frequency interactions. A similar interaction-driven behavior is observed for oscillation amplitude. Although amplitude alone did not exhibit a significant main effect, its influence becomes pronounced at higher frequencies and depends strongly on fin morphology. In general, thrust increases approximately linearly with amplitude at moderate-to-high frequencies, whereas at low frequencies its effect is minimal. This trend is evident in Figure~\ref{fig:fin-thrust-results}a, where amplitude primarily affects the slope of the frequency response curves. Notably, the eel fin shows negligible sensitivity to amplitude, in contrast to all other morphologies. Material and thickness effects are best understood through their interaction with both fin morphology and structural compliance, rather than as independent factors. Across most fin types, the rigid and semi-soft configurations exhibit broadly similar thrust responses, with only minor deviations in magnitude and slope across frequencies (Figure~\ref{fig:fin-thrust-results}a). In these cases, the thickness profile has a negligible influence on performance, indicating that when the structure remains relatively stiff, geometric tapering does not substantially alter the hydrodynamic response. An exception to this trend is observed for fins with larger effective surface areas, such as the boxfish, where thickness variations introduce more noticeable differences in thrust, likely due to increased fluid–structure interaction over the fin surface. In contrast, the soft material configuration displays a fundamentally different behavior. Here, the thickness profile plays a critical role, with the conical profile consistently outperforming the constant thickness profile across all fin types and frequencies. This effect is particularly evident in the trout, mackerel, tuna, and average fins, where the conical soft fins not only achieve higher peak thrust but also exhibit steeper increases in thrust with frequency. The increased performance of the conical profile suggests that spanwise stiffness gradients enhance the deformation pattern of the fin, enabling more favorable kinematics and improved momentum transfer to the fluid. This distinction highlights that for sufficiently compliant structures, geometric distribution of material—rather than just overall stiffness—becomes a key design parameter. Overall, these results demonstrate that material and thickness effects are negligible in stiff regimes but become dominant in compliant regimes, where they strongly influence propulsion efficiency and thrust generation.

The optimal configuration for each fin type (Table~\ref{tab:bestfins}) was identified using Bonferroni-adjusted confidence intervals. The corresponding thrust curves are shown in Figure~\ref{fig:fin-thrust-results}b. At low frequencies, the boxfish fin produces the highest thrust; however, a transition occurs around 3.5–4~TB~s$^{-1}$, beyond which the trout and average fins outperform it. At higher frequencies, the average fin achieves the overall highest thrust. This trend is summarized in Figure~\ref{fig:fin-thrust-results}c, which compares the maximum thrust attained by each fin at its optimal operating condition.

\begin{table}[htbp]
\centering
\caption{The best caudal fin configuration for each of the bio-inspired caudal fin designs.}
\label{tab:bestfins}
\begin{tabular}{lccccc}
\hline
Fin type & Material & Thickness & Amplitude [BL] & Frequency [TB s$^{-1}$] & Max Thrust [N] \\
\hline

Eel & Semi-Soft & Constant & 0.21 & 7 & 0.0016 \\

Trout & Soft & Conical & 0.21 & 7 & 0.242 \\

Mackerel & Soft & Conical & 0.21 & 7 & 0.152 \\

Tuna & Rigid & Conical & 0.21 & 6.5 & 0.132 \\

Boxfish & Semi-Soft & Conical & 0.21 & 2.5 & 0.152 \\

\textbf{Average} & \textbf{Soft} & \textbf{Conical} & \textbf{0.21} & \textbf{7} & \textbf{0.256} \\

\hline
\end{tabular}
\end{table}
\subsection{Numerical Body-Drag Characterization}
\label{sec:body-results}

\begin{figure}[!htbp]
    \centering
    \includegraphics[width=\linewidth]{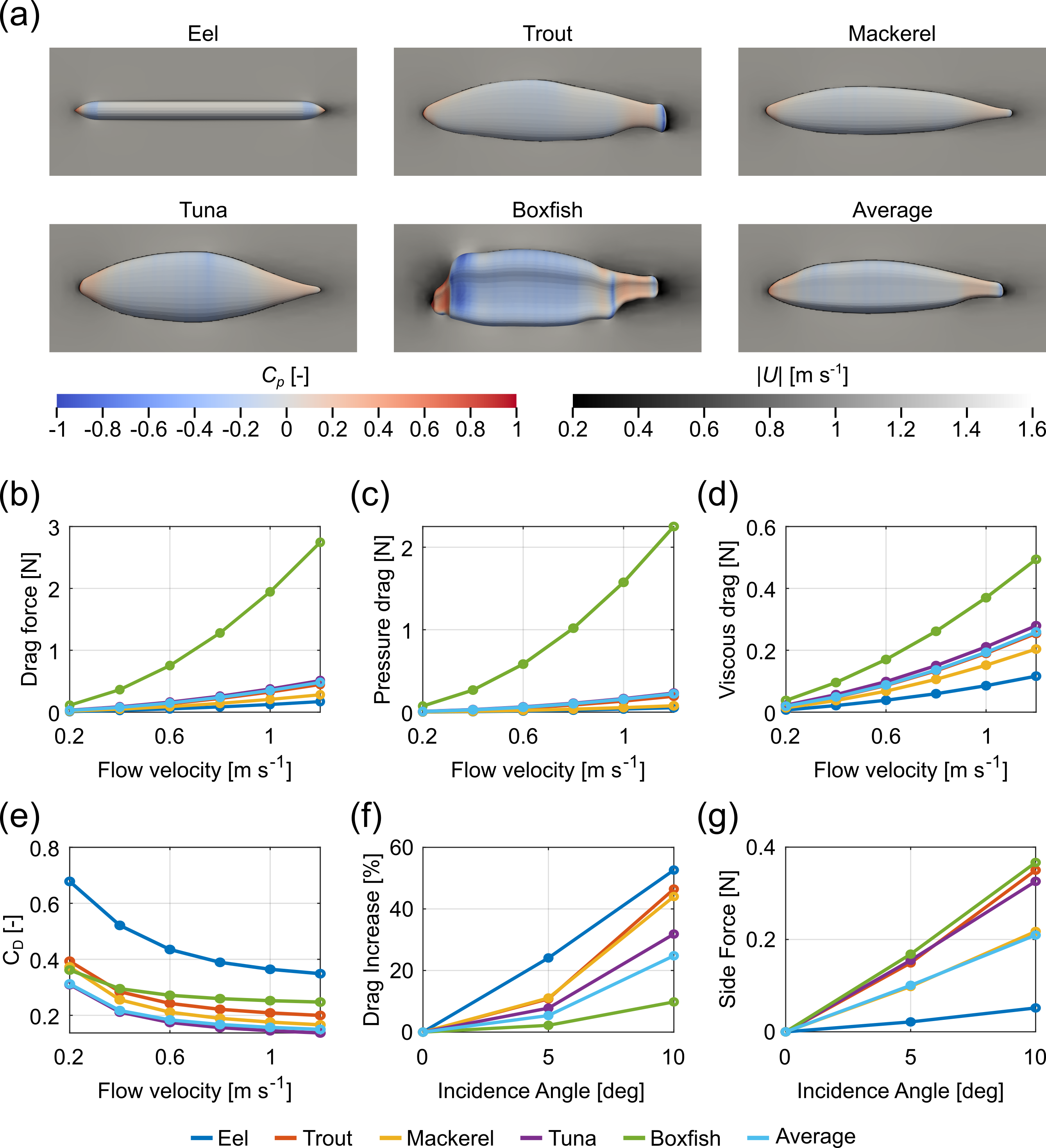}
    \caption{Summary of numerical body-form drag simulation results. (a) Surface colored by pressure coefficient $C_p$, with velocity magnitude $|U|$ shown in the surrounding flow field. (b) Drag force acting on each body over the investigated flow regime. (c)--(d). Pressure and viscous drag components. (e) Drag coefficients for the six body shapes. (f) Drag increase in percentage when bodies are experiencing off-axis flow at a velocity of 0.6~m~s$^{-1}$. (g) Side force resulting from off-axis flow.}
    \label{fig:drag-results}
\end{figure}

After identifying the caudal fin configuration for the robotic platform, the hydrodynamic performance of the candidate body geometries was evaluated using the CFD procedure described in Section~\ref{sec:sim}. Figure~\ref{fig:drag-results}a shows the surface pressure coefficient together with the surrounding velocity magnitude field for each body at $U_\infty = 1.2$~m~s$^{-1}$.

The boxfish body exhibited the strongest pressure gradients and the largest disturbance to the surrounding flow field, indicating a poor streamlining. In contrast, the eel, trout, mackerel, tuna, and averaged bodies showed relatively similar pressure distributions, with a stagnation region near the snout, near-zero or mildly negative pressure over much of the body, and a localized pressure recovery near the caudal peduncle. The corresponding velocity fields showed relatively smooth acceleration around the streamlined bodies, followed by localized deceleration in the posterior region.

The numerical drag results for the six body geometries are summarized in Figure~\ref{fig:drag-results}b--e, and confirm the indications from the surface pressures and velocity magnitude fields in Figure~\ref{fig:drag-results}a. As expected, the boxfish body experiences the largest absolute drag over the full investigated flow range, whereas the eel body experiences the lowest drag (Figure~\ref{fig:drag-results}b). These two bodies represent opposite extremes in body form, with the eel being long and slender and the boxfish being deep and bluff. Among the remaining bodies, the mackerel exhibits the second-lowest drag, which is consistent with its relatively shallow and streamlined profile. The tuna and average bodies perform very similarly, with the average body showing slightly lower absolute drag than the tuna over much of the investigated range. 
The decomposition of drag into pressure and viscous contributions is shown in Figure~\ref{fig:drag-results}c--d. The boxfish is clearly dominated by pressure drag, indicating a substantial pressure imbalance between the front and rear of the body, which is characteristic of a poorly streamlined shape. In contrast, the eel, trout, mackerel, tuna, and average bodies are all dominated by viscous drag, while maintaining comparatively low pressure drag. This indicates that these bodies generate less large-scale pressure loss and wake-related drag, consistent with more streamlined flow development and reduced flow separation.
The drag coefficients are shown in Figure~\ref{fig:drag-results}e. Rather than remaining constant, the drag coefficients decrease with increasing flow velocity for all bodies over the simulated range, indicating Reynolds-number dependence within the investigated regime. The eel exhibits the largest drag coefficient throughout, despite having the lowest absolute drag force, reflecting the fact that its highly elongated body is not particularly efficient when referenced against its frontal area. The boxfish also shows comparatively high drag coefficients, though still markedly lower than the eel. The trout and mackerel show progressively lower drag coefficients, consistent with their increasingly streamlined body forms. Again, the tuna and average bodies follow each other closely, with only minor differences across the tested speeds.

When the bodies are subject to off-axis flows their drag forces increase, and as expected, the drag increase grows with increasing incidence angle for all bodies (see Figure~\ref{fig:drag-results}f). The trout and mackerel bodies display similar off-axis sensitivity, with moderate increases in drag at both angles. The tuna body is less sensitive than these two shapes, but still experiences a noticeable drag penalty under misalignment. The average body consistently shows lower drag increase than the trout, mackerel, and tuna, indicating a lower sensitivity to off-axis flow while still maintaining favorable aligned-flow performance. This supports the interpretation of the average body as a balanced compromise geometry. Unexpectedly, the eel exhibited the largest relative drag increases for both the incidence angles, while the boxfish exhibited the smallest. This should not be interpreted as inferior or superior hydrodynamic performance under off-axis conditions, but rather as a consequence of their already small and large aligned drag, respectively. In other words, the eel remains the best-performing body in terms of absolute drag and boxfish the worst-performing. Even a small increase in drag for the eel body will be a large relative increase and the boxfish's poor baseline performance leaves less room for further relative deterioration. The side-force results further supported these trends (Figure~\ref{fig:drag-results}g). The eel produces the lowest side force at both incidence angles, while the mackerel and average bodies also perform favorably. In contrast, the trout, tuna, and boxfish produce larger side forces under off-axis flow. Taken together, these numerical results indicate that the average body provides a balanced trade-off between low aligned-flow drag, reduced off-axis drag sensitivity, and moderate side-force generation. The averaged geometry was therefore selected as a suitable body design for a robotic fish intended to span multiple BCF locomotion modes. The drag forces, drag coefficients, and off-axis flow metrics at the maximum flow velocity are summarized in Table~\ref{tab:drag}.

\begin{table}[htbp]
\centering
\caption{Drag results from simulation at the maximum velocity, including off-axis performance at 10$^{\circ}$.}
\label{tab:drag}
\begin{tabular}{lccccc}
\hline
Body & Drag [N] & Pressure/Viscous drag Dist. & $C_D$ [-] & Drag increase [\%] & Side force [N]\\
\hline

Eel & 0.17 & 31.84/68.16 & 0.349 & 52.62 & 0.052 \\

Trout & 0.45 & 43.16/56.84 & 0.200 & 46.45 & 0.350 \\

Mackerel & 0.28 & 27.47/72.53 & 0.166 & 44.03 & 0.217 \\

Tuna & 0.51 & 45.26/54.74 & 0.137 & 31.84 & 0.326 \\

Boxfish & 2.75 & 82.01/17.99 & 0.247 & 9.81 & 0.367 \\

\textbf{Average} & \textbf{0.48} & \textbf{46.22/53.78} & \textbf{0.150} & \textbf{24.79} & \textbf{0.210} \\

\hline
\end{tabular}
\end{table}
\subsection{Robot Implementation and Mechanical Mode Switching}
Based on the morphometric, caudal-fin, and body-drag results presented above, the final \textit{OmniFish} architecture was implemented as described in this section. The robot design is presented here as an outcome of the preceding design-selection process. The average body showed favorable drag characteristics comparable to the tuna-inspired morphology while exhibiting reduced sensitivity to off-axis flow conditions expected during BCF locomotion. Similarly, the semi-soft conical average caudal fin produced the most favorable thrust response among the tested fin configurations and was therefore selected.

\begin{figure}[htbp]
    \centering
    \includegraphics[width=\linewidth]{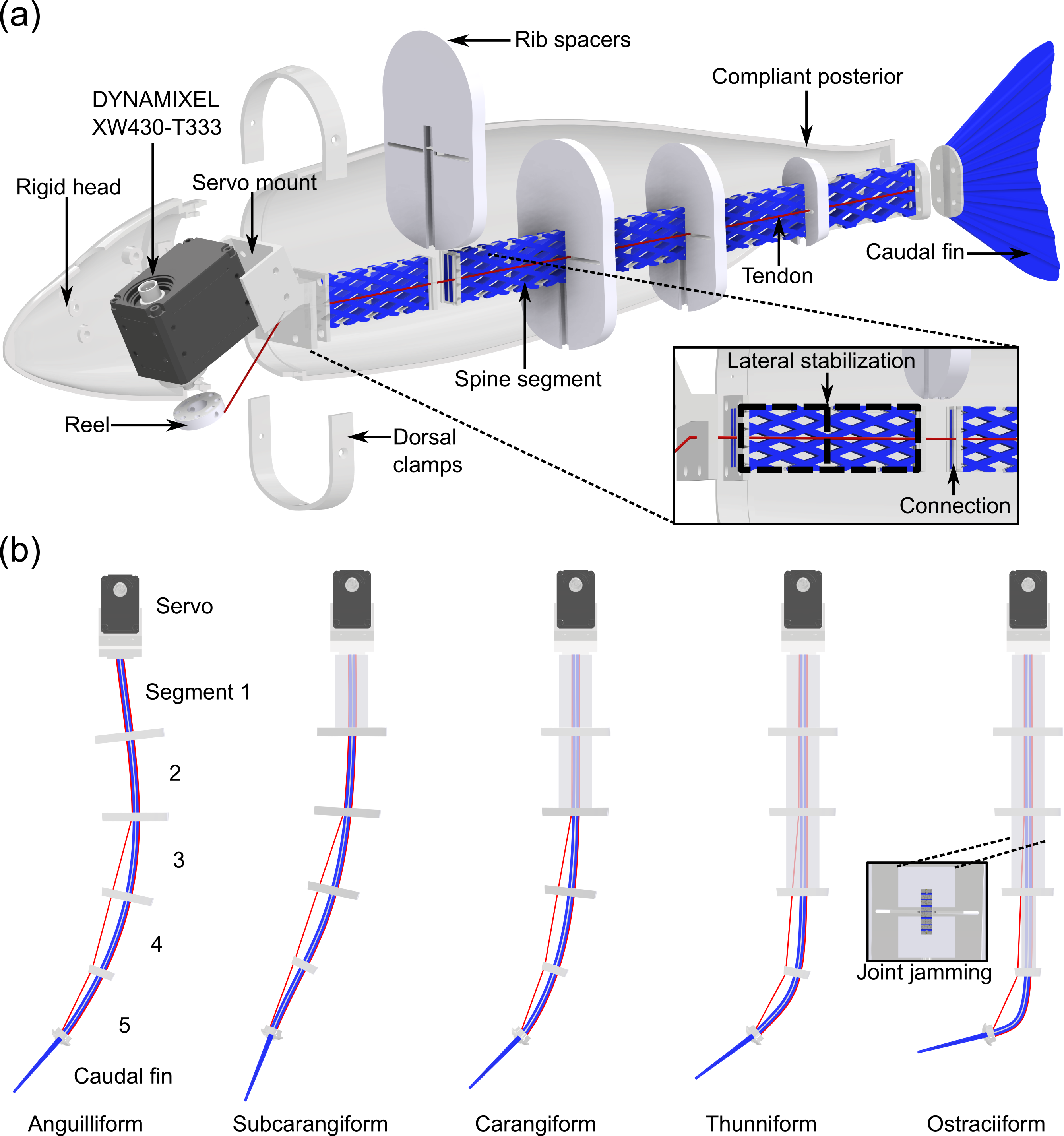}
    \caption{Fish inspired robot design. (a) Exploded view of the robot design with highlighted components. (b) Spine deformation for the five investigated swimming modes with zoomed cross sectional view of the joint jamming mechanism.}
    \label{fig:robotdesign}
\end{figure}

\subsubsection*{Body architecture}

\textit{OmniFish} was designed with a standard length of 40~cm with the caudal fin extending an additional 5.42~cm back, giving a full robot length of 45.42~cm (see Supplementary information S8). The body was divided into a rigid anterior head section spanning the first 10~cm from the snout and a compliant posterior section spanning the remaining 30~cm, as illustrated in Figure~\ref{fig:robotdesign}a. This distribution was selected to emulate the full range of BCF swimming modes, where anguilliform locomotion requires distributed body deformation while progressively more posterior actuation characterizes the remaining modes. The rigid head section housed the actuation system consisting of a DYNAMIXEL XW430-T333 servo motor (ROBOTIS Co., Ltd.), servo mount, and cable communication interface. The servo mount was rigidly fixed within the head and connected directly to the compliant spine extending through the posterior body. Four rib spacers distributed along the compliant section supported the external body skin and transmitted deformation to the body surface.

\subsubsection*{Compliant spine and body construction}

The spine consisted of five modular segments numbered anterior to posterior (Figure~\ref{fig:robotdesign}b). Five segments were selected to enable progressive locking corresponding to the five BCF-inspired locomotion modes. Each segment incorporated a compliant lattice structure enabling controlled deformation and a central longitudinal stabilizing frame to constrain deformation primarily to the horizontal plane while reducing torsional deformation and out-of-plane buckling (see more in Supplementary information S8). Adjacent segments interfaced through press-fit connections integrated into rigid PETG rib spacers. The rib spacers overlapped neighboring segment interfaces, forming rigid junctions while preserving compliance within the segment bodies. Integrated slots within the ribs provided tendon routing paths. The posterior body was formed from 2~mm thick EcoFlex 00-30 silicone (Smooth-On, Inc.) cast using a dedicated 3D printed mold that can be seen in Supplementary information S9. The skin was fitted over the spine assembly and clamped between the rigid head and modular dorsal and ventral clamps. Posteriorly, the caudal fin was attached through the skin using screws threading directly into the terminal spine segment. This formed a modular sealing mechanism allowing replacement of the compliant skin while incorporating an integrated gasket between the rigid and compliant sections.

\subsubsection* {Antagonistic tendon-driven actuation}

Actuation was achieved using two antagonistic 0.1~mm nylon tendons routed symmetrically along the spine. The tendons originated at a reel mounted directly to the Dynamixel output shaft and terminated at the posterior spine segment adjacent to the caudal fin. Tendons were manually tensioned to introduce slight compression while avoiding visible static curvature. The actuation mechanism was inspired by the variable thickness flexible beam approach proposed by Ta~\cite{ta2024single}, where compliant beams generate S-shaped body deformations. In the present work significant pre-tension was intentionally omitted to avoid static bias and preserve mode switching capability. Instead, S-shaped body profiles emerged dynamically through rapid alternating actuation, where inertial effects generated transient counter-curvatures during direction reversal. Experimental servo excursions ranged from 10--78$^\circ$, corresponding to approximately 2--14~cm peak-to-peak tail amplitudes. The mapping between servo rotation and tail amplitude is provided in Supplementary Information S6.

\subsubsection*{Mechanical generation of BCF-inspired swimming modes}

Swimming modes were generated by mechanically jamming spine segments (Figure~\ref{fig:robotdesign}b). Anguilliform locomotion was achieved without locked segments, while Subcarangiform, Carangiform, Thunniform, and Ostraciiform modes were generated by progressively locking segments 1, 1--2, 1--3, and 1--4, respectively. This progressively shifted deformation posteriorly while increasing effective body stiffness. Segment locking was achieved using removable PETG jamming blocks inserted between rib interfaces. The jamming blocks enclosed the spine segment and prevented local deformation, fully constraining bending. Mode transitions required disassembly of the compliant posterior body. The caudal fin, dorsal and ventral clamps, and silicone skin were removed to expose the spine, after which jamming blocks were inserted and tendon tension re-established before reassembly.

\subsubsection*{Buoyancy compensation}

Although the body could theoretically be fully sealed, the robot was intentionally flooded during operation to regulate buoyancy. As the inserted jamming blocks displaced internal water volume and introduced positive buoyancy, zinc ballast masses were added for compensation. Anguilliform locomotion used 5~g ballast at rib two, Subcarangiform used 15~g distributed between ribs two and three, Carangiform used 30~g, while Thunniform and Ostraciiform used 45~g distributed along the jamming assembly of segment three. 
Because the ballast mass and placement varied between swimming modes, their potential influence on the hydrostatic and rigid-body inertial properties was evaluated separately, see Supplementary information S10. The analysis showed only small configuration-dependent variations, indicating that the ballast primarily compensated for the buoyancy changes introduced by the PETG jammers. Across configurations, the robot had a mean total volume of 1.019~L, a dry mass of 0.343~kg, and a mean flooded mass of 1.065~kg, resulting in a slightly negatively buoyant configuration.
\subsection{Kinematics of the Five BCF-Inspired Modes}
\begin{figure}[htbp]
    \centering
    \includegraphics[width=\linewidth]{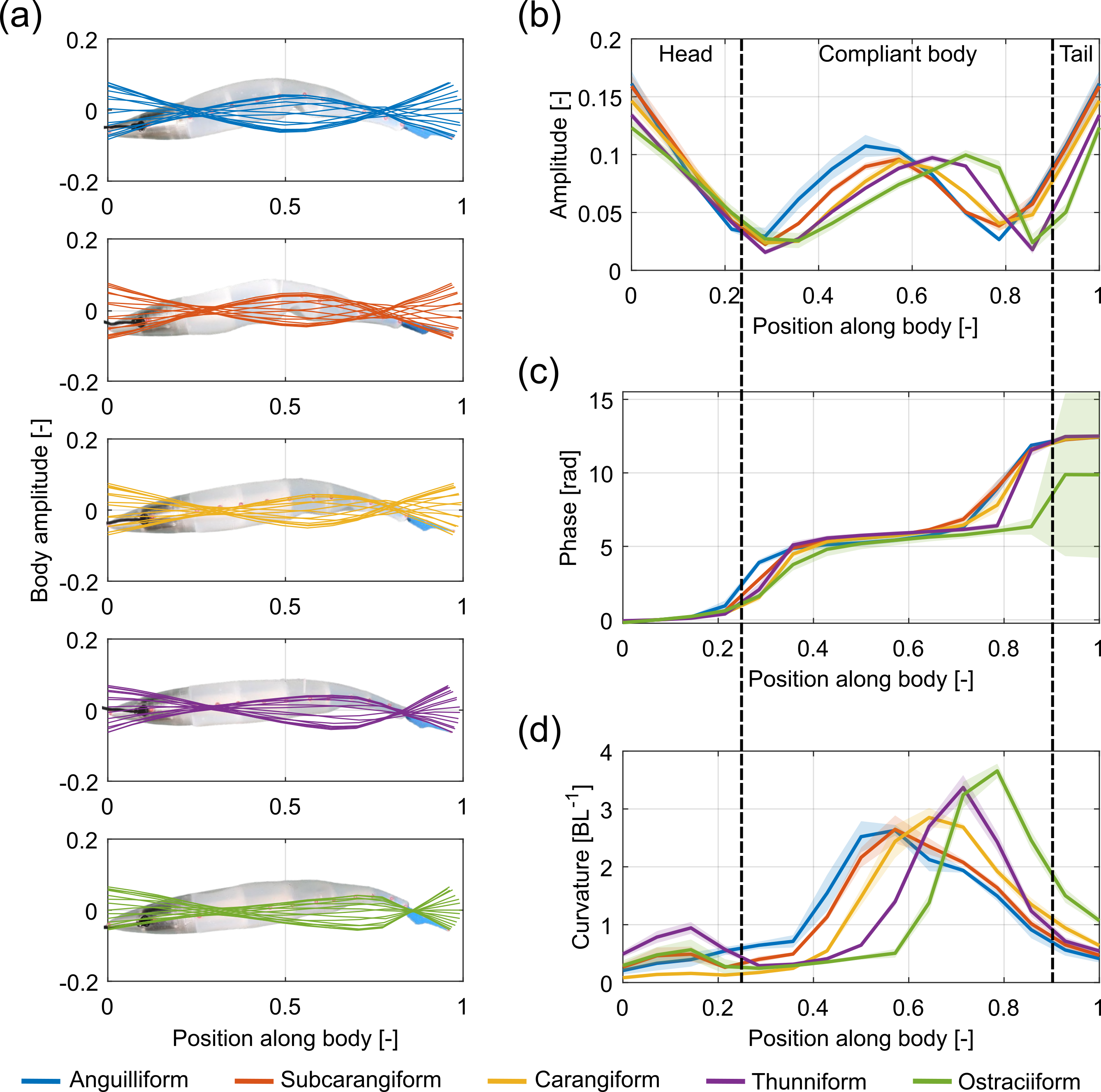}
    \caption{Kinematic analysis of the five BCF-inspired configurations at 1~TB~s$^{-1}$. (a) Reconstructed midlines over one representative cycle, illustrating the spatial distribution of body deformation. (b) Normalized peak-to-peak amplitude envelopes along the body; shaded regions indicate cycle-to-cycle variation. (c) Relative phase progression along the propulsive structure. (d) Mean absolute curvature profiles along the body.}
    \label{fig:kinematics-results}
\end{figure}

The implemented \textit{OmniFish} platform was evaluated to determine whether progressive segment jamming could generate distinct BCF-inspired deformation patterns and how the distribution of body deformation changed across stiffness configurations. Figure~\ref{fig:kinematics-results} summarizes the reconstructed midlines, normalized amplitude envelopes, relative phase progression, and mean absolute curvature.

Mode-dependent kinematics were assessed using the procedure described in Section~\ref{sec:kin} (watch Supplementary video S2). Figure~\ref{fig:kinematics-results}a shows reconstructed midlines for the five investigated configurations at 1~TB~s$^{-1}$. The midlines reveal appreciable anterior motion and two recurring low-amplitude regions near the anterior body and caudal-peduncle region for all modes. These features are not characteristic of the smooth midline envelopes commonly reported for biological fish and may arise from the discrete segmented architecture, structural transitions along the spine, residual rigid-body motion, or interaction with the tethered experimental setup. Their individual contributions could not be separated in the present analysis.
Despite these non-biological features, the midlines show a progressive posterior localization of deformation as additional anterior spine segments are jammed. The Anguilliform configuration exhibits deformation over a comparatively large proportion of the body, whereas the Subcarangiform and Carangiform configurations increasingly concentrate motion in the posterior body. The Thunniform and Ostraciiform configurations maintain a largely rigid anterior body and exhibit predominantly localized caudal motion.
The normalized amplitude envelopes in Figure~\ref{fig:kinematics-results}b did not follow the smooth posteriorly increasing envelopes commonly reported for biological fish. Instead, all configurations exhibited local amplitude minima. The first minimum occurred near the transition between the rigid head and compliant spine. In the more compliant configurations, a second minimum occurred near the posterior body, whereas in the Thunniform and Ostraciiform configurations this minimum shifted further toward the caudal region. These local minima indicate that deformation was transmitted non-uniformly through the segmented spine rather than as a continuous biological body wave. Figure~\ref{fig:kinematics-results}c shows the relative phase along the propulsive structure. Phase generally increased toward the caudal region for all configurations, indicating posterior progression of deformation. However, the increase was spatially non-uniform and contained localized changes in slope. These changes occurred near regions of low oscillation amplitude, where phase estimates are inherently less reliable because the lateral motion approaches zero. The phase results therefore support an overall posterior progression of deformation, but not a smooth traveling wave comparable to that of biological fish. The mean absolute curvature profiles in Figure~\ref{fig:kinematics-results}d further illustrate the effect of progressive segment jamming. Curvature remained low within the rigid anterior section and increased within the compliant body. As more anterior segments were jammed, the region of maximum curvature shifted posteriorly, demonstrating a transition from distributed body bending toward localized caudal deformation. Curvature decreased near the caudal tip for all modes, which may reflect the comparatively high stiffness of the caudal fin as well as reduced accuracy of curvature estimation near the endpoint. 

Table~\ref{tab:kinematic-summary} summarizes the kinematic analysis. Overall, the kinematic analysis shows that \textit{OmniFish} reproduces the broad posterior localization of deformation associated with the BCF spectrum, while differing from biological fish in the smoothness and continuity of amplitude, phase, and curvature distributions.

\begin{table}[!htbp]

\centering

\caption{Summary of kinematic metrics for the five implemented BCF-inspired swimming modes. Amplitudes are reported as peak-to-peak values normalized by body length (BL) and curvature is expressed in BL$^{-1}$. Values are reported as mean $\pm$ SD across complete tail-beat cycles.}

\begin{tabular}{l c c c c c}

\hline

& Anguilliform & Subcarangiform & Carangiform & Thunniform & Ostraciiform \\

\hline

Head amplitude [-]
& $0.096 \pm 0.004$
& $0.100 \pm 0.006$
& $0.098 \pm 0.006$
& $0.088 \pm 0.002$
& $0.088 \pm 0.006$ \\

Tail amplitude [-]
& $0.134 \pm 0.010$
& $0.132 \pm 0.010$
& $0.122 \pm 0.006$
& $0.104 \pm 0.002$
& $0.086 \pm 0.006$ \\

Head:tail amplitude ratio [-]
& $0.713 \pm 0.018$
& $0.759 \pm 0.014$
& $0.804 \pm 0.007$
& $0.841 \pm 0.022$
& $1.007 \pm 0.016$ \\

Max amplitude [-]
& $0.162 \pm 0.012$
& $0.160 \pm 0.010$
& $0.146 \pm 0.006$
& $0.134 \pm 0.004$
& $0.124 \pm 0.006$ \\

Max curvature, $\kappa_{\max}$ [BL$^{-1}$]
& $2.003 \pm 0.163$
& $2.072 \pm 0.190$
& $2.169 \pm 0.099$
& $2.645 \pm 0.145$
& $3.262 \pm 0.092$ \\

Location of max curvature [-]
& $0.543 \pm 0.039$
& $0.571 \pm 0.000$
& $0.643 \pm 0.000$
& $0.714 \pm 0.000$
& $0.786 \pm 0.000$ \\

\hline

\end{tabular}

\label{tab:kinematic-summary}

\end{table}

\subsection{Swimming Performance of the Five BCF-Inspired Modes}

Swimming experiments were performed for all modes over the frequency range 0.6, 1.0, 1.5, and 2.0~TB~s$^{-1}$ as described in Section~\ref{sec:swim}, with representative snapshots shown in Figure~\ref{fig:swimming-results}a (swimming performance can be observed in Supplementary video S3). Swimming velocity and COT calculated over the identified steady-swimming intervals, following the procedure described in Section~\ref{sec:swim} and illustrated in Supplementary Information S7, were analyzed using a two-way ANOVA with swimming mode and tail-beat frequency as factors. Significant main effects of both mode and frequency were found for velocity ($F=1874.83$, $p<0.001$ and $F=439.1$, $p<0.001$, respectively) and COT ($F=97.07$, $p<0.001$ and $F=130.34$, $p<0.001$, respectively). Significant interaction effects were also observed for velocity ($F=308.05$, $p<0.001$) and COT ($F=30.90$, $p<0.001$), indicating that locomotion performance depended strongly on both swimming mode and operating frequency. A complete statistical analysis can be found in Supplementary information S11.

Anguilliform, Subcarangiform, and Carangiform locomotion exhibited similar performance with respect to both swimming velocity and COT (Figure~\ref{fig:swimming-results}b-c). All three modes showed peak performance near 1~TB~s$^{-1}$, indicating a common optimum operating frequency. At lower frequencies deformation developed slowly and primarily produced C-shaped body profiles, while the transient counter-curvatures associated with inertial S-shaped motion became increasingly apparent near and above 1~TB~s$^{-1}$. However, at higher frequencies tail amplitude progressively degraded, resulting in reduced swimming performance. In contrast, Thunniform and Ostraciiform locomotion improved with increasing frequency and reached peak performance near 1.5~TB~s$^{-1}$. Ostraciiform remained relatively insensitive between 1.5 and 2.0~TB~s$^{-1}$, whereas Thunniform exhibited a stronger decline due to earlier degradation of tail amplitude. Consequently, 1.5~TB~s$^{-1}$ was selected as the representative optimum frequency for both modes. The optimum operating frequency shifted toward higher tail-beat frequencies as body rigidity increased. This trend qualitatively resembles stiffer biological swimmers, which commonly operate at higher frequencies. These frequencies represent the experimentally observed performance optima and should not be interpreted as measured structural resonance frequencies.

\begin{figure}[htbp]
    \centering
    \includegraphics[width=\linewidth]{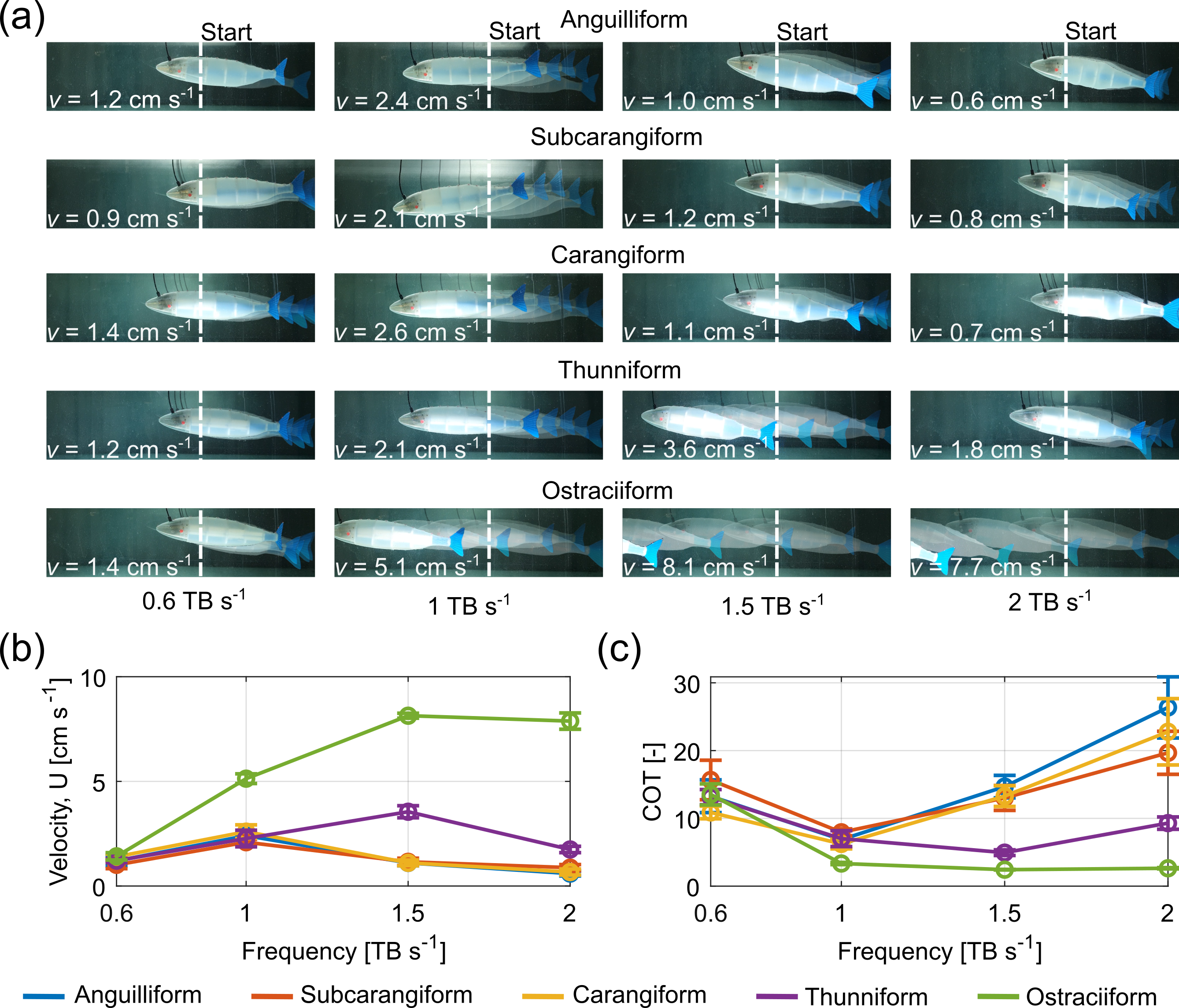}
    \caption{Robot swimming results. (a) Experimental snapshots over an 8~s period. Transparency gradients illustrate the displacement over time. (b) Swimming velocity for the five investigated modes over the tested frequency range. (c) Cost of transport (COT) over the tested frequency range.}
    \label{fig:swimming-results}
\end{figure}

Optimum swimming performance is summarized in Table~\ref{tab:bestswims}. Tail amplitude degraded more rapidly with increasing frequency in more compliant body configurations (Supplementary Information S6), likely contributing to the shift in optimum operating frequency toward higher values with increasing body rigidity. Ostraciiform locomotion achieved the highest swimming performance and efficiency, followed by Thunniform locomotion, suggesting that the present architecture favors localized tail propulsion. Although the remaining modes exhibited lower straight-line performance, they may provide maneuverability advantages not evaluated in the present study.

\begin{table}[htbp]
\centering
\caption{Best swimming performance of the five BCF modes summarized as mean swimming velocity and (COT) $\pm$ standard deviation at the best-performing tested tail-beat frequency.}
\label{tab:bestswims}
\begin{tabular}{lcccc}
\hline
Mode & Frequency [TB s$^{-1}$] & Velocity [cm s$^{-1}$] & Velocity [BL s$^{-1}$] & COT [-]\\
\hline

Anguilliform & 1 & 2.40 $\pm$ 0.08 & 0.053 $\pm$ 0.002 & 6.86 $\pm$ 0.24 \\

Subcarangiform & 1 & 2.09 $\pm$ 0.14 & 0.046 $\pm$ 0.003 & 7.94 $\pm$ 0.50 \\

Carangiform & 1 & 2.60 $\pm$ 0.33 & 0.057 $\pm$ 0.007 & 6.23 $\pm$ 0.73 \\

Thunniform & 1.5 & 3.55 $\pm$ 0.29 & 0.078 $\pm$ 0.006 & 4.94 $\pm$ 0.42 \\

\textbf{Ostraciiform} & 1.5 & \textbf{8.14 $\pm$ 0.12} & \textbf{0.179 $\pm$ 0.003} & \textbf{2.42 $\pm$ 0.04} \\

\hline
\end{tabular}
\end{table}

Across all swimming experiments the robot operated at Reynolds numbers ranging approximately from $2.5\times10^3$ to $4.3\times10^4$ and Strouhal numbers ranging from approximately 1.38 to 10.24. These dimensionless values were calculated to characterize the experimental operating regime rather than to serve as primary performance metrics. The robot operated at lower Reynolds numbers and notably higher Strouhal numbers than typically reported for efficient biological cruising, primarily due to the low forward swimming velocity and tethered experimental configuration.

\section{Discussion}
\subsection{Morphological Averaging as a Design Strategy}

This work explored morphological averaging as a design strategy for developing a robotic fish capable of spanning the full range of BCF locomotion modes. Rather than optimizing toward a single biological archetype, the objective was to combine representative morphological features from multiple species into a unified body and caudal fin geometry suitable for multimodal swimming. This approach produced an averaged morphology that retained several features associated with streamlined BCF swimmers, while also incorporating geometric characteristics from less streamlined but mechanically distinct forms such as the boxfish. The pointwise arithmetic average was not assumed to be hydrodynamically optimal, nor was it intended to outperform every species-specific morphology. Instead, it provided an equal-weight geometric reference that avoided selecting one species-specific body as the sole morphological baseline for all five stiffness configurations. This was appropriate for the present study because the primary objective was to investigate whether changes in stiffness distribution could shift deformation from whole-body undulation to localized tail oscillation within a single shared morphology.

The caudal fin characterization showed that the averaged fin design provided a favorable thrust response among the investigated fin morphologies. This suggests that an averaged fin geometry can capture useful propulsive features from multiple biological templates rather than relying on a single species-specific design. Similarly, the CFD analysis showed that the averaged body provided a balanced hydrodynamic compromise. Although it did not minimize every drag-related metric, it combined favorable aligned-flow drag characteristics with reduced sensitivity to off-axis flow compared with several of the individual species-inspired bodies. This is particularly relevant for multimodal robotic fish, where the body is expected to experience different deformation patterns and local flow orientations across swimming modes. A body optimized exclusively for low drag in aligned flow may not provide the most favorable response during undulatory swimming, where sections of the body experience changing local incidence angles and lateral loading. The averaged morphology was therefore selected as a compromise across these competing hydrodynamic requirements rather than as a single-metric optimum.

These findings suggest that morphological averaging can provide a useful design framework when the goal is versatility rather than specialization. In contrast to bioinspired designs that reproduce one specific fish morphology for one specific locomotion mode, the present approach uses biological diversity as a design envelope. This may be valuable for future robotic fish intended to operate across different swimming regimes, environments, or mission requirements. However, morphological averaging should not be interpreted as a universal optimization strategy. Its effectiveness depends on the selected biological templates, the weighting of morphological features, and the intended performance objectives. The present CFD comparison evaluates the rigid-body hydrodynamic characteristics of the averaged and species-specific geometries, but it does not establish that the averaged morphology would outperform a trout-, mackerel-, or other species-inspired body when each is combined with the same five stiffness configurations. Demonstrating such superiority would require otherwise equivalent robotic platforms with different external morphologies to be evaluated under identical actuation and stiffness conditions. Future work could therefore explore direct comparisons with non-averaged control morphologies, as well as weighted or performance-guided averaging, where species contributions are adjusted according to desired metrics such as thrust, drag, maneuverability, stability, or efficiency.

\subsection{Mechanical Mode Switching and Body Stiffness Distribution}
\label{sec:disc-mech}

The quantitative kinematic analysis confirmed the intended posterior localization of deformation but also revealed local kinematic nodes and non-uniform phase progression associated with the discrete segmented architecture, indicating that the generated modes are BCF-inspired rather than exact reproductions of biological body waves. Similar concepts have previously been demonstrated using actively jammed tensegrity structures~\cite{wang2026adaptive}; however, the present work incorporates all five BCF locomotion modes through passive segment locking within a single-actuator tendon-driven architecture. By progressive jamming anterior spine segments, \textit{OmniFish} shifted deformation posteriorly and generated kinematic patterns resembling Anguilliform, Subcarangiform, Carangiform, Thunniform, and Ostraciiform locomotion. Increasing rigidity generally resulted in increased swimming speed and reduced COT, suggesting that reduced body deformation losses enabled more effective transfer of actuation energy toward caudal propulsion. This confirms that body stiffness distribution is a key design variable for generating distinct BCF swimming modes.

One of the most unexpected findings was the performance of the Ostraciiform configuration. Biological Ostraciiform swimmers are generally characterized by relatively slow locomotion; however, within \textit{OmniFish} this mode achieved both the highest swimming velocity and the lowest COT. Initially, Thunniform locomotion was expected to provide the best compromise between speed and efficiency, consistent with biological tuna locomotion. Instead, performance improved progressively with increasing body rigidity, suggesting that the present architecture mechanically favors localized tail propulsion. Several factors may explain this behavior. First, the robot relies on a single actuator driving two antagonistic tendons. In highly complaint configurations, force transmission through the body is more sensitivity to tendon pretension, whereas progressively rigid configurations inherently maintain higher transmission efficiency. 
Second, the compliant posterior body requires substantial energy for deformation, causing a larger fraction of the input energy to be dissipated internally rather than converted into useful thrust. Furthermore, body deformation is transmitted through discrete rib interfaces between the spine and compliant skin. This discontinuous coupling may reduce deformation transmission efficiency in highly undulatory modes. The kinematic analysis further highlights the influence of the segmented mechanical architecture on body-wave propagation. In contrast to the smooth posteriorly increasing amplitude envelopes typically reported for biological swimmers, \textit{OmniFish} exhibited local amplitude minima and comparatively large anterior motion. This suggests that part of the imposed motion is spent moving the anterior body rather than being progressively amplified toward the caudal region, which may reduce the effectiveness of forward propulsion. The relative phase nevertheless increased toward the tail, indicating posterior propagation of deformation, but the progression was spatially non-uniform and became less reliable near the low-amplitude regions. Together, these results indicate that the robot generates a posteriorly propagating deformation pattern, but not a smooth continuous traveling wave comparable to biological fish. The discrete spine segments, rib interfaces, and transitions between rigid and compliant regions likely contribute to these kinematic discontinuities. A more continuously distributed stiffness architecture and improved coupling between the spine and skin may therefore enable smoother wave propagation and reduce unnecessary anterior motion. The posterior shift in the region of maximum curvature nevertheless confirms that progressive jamming successfully relocates the primary bending region toward the caudal body. The Ostraciiform configuration also retained a small amount of posterior body deformation rather than producing purely localized tail oscillation. Consequently, its kinematics partially resembled localized Thunniform propulsion. Future iterations could further isolate tail oscillations by redistributing segment lengths, for example by shortening the terminal segment while increasing the lengths of the anterior segments.

The low Reynolds and high Strouhal operating regime limits direct comparison with the hydrodynamic performance of biological fish. However, the relevance of the implemented modes in the present study lies primarily in their distinct spatial distributions of body deformation, rather than in exact dynamic similarity to the corresponding biological swimmers. The common platform provides a controlled means of investigating how stiffness localization changes kinematics and straight-line performance while morphology and actuation remain fixed. Whether the more compliant modes provide their expected functional advantages in maneuvering, disturbance rejection, or confined environments remains to be established under less constrained and more biologically comparable operating conditions.

\subsection{Comparison with State-of-the-Art}

\begin{figure}[htbp]
    \centering
    \includegraphics[width=\linewidth]{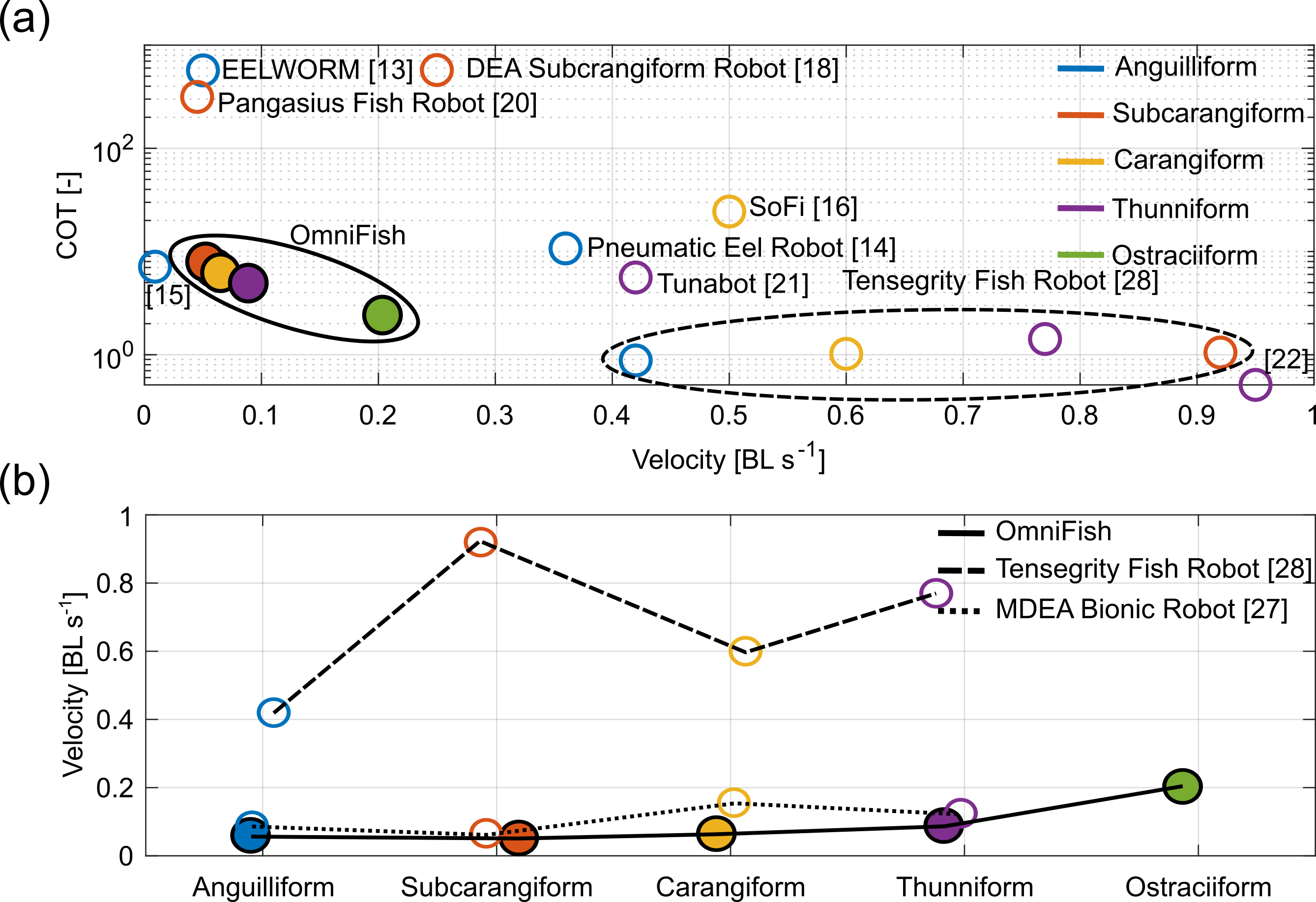}
    \caption{Comparison with the SOTA, at a comparable actuation frequency. \textit{OmniFish} is shown as colored markers with black edging and the SOTA as hollow circles, colors represent primary locomotion mode. (a) Velocity VS. COT of all the robots from previously mention SOTA, with available metrics for efficiency measure. The COT axis employs logarithmic scale, as some samples have very large values. (b) Direct comparison with other multi-BCF-mode robot with respect to speed.}
    \label{fig:sota-comp}
\end{figure}

Comparison with state-of-the-art robotic fish platforms is challenging, as many studies do not report efficiency measures such as COT or provide sufficient information for reconstruction. Nevertheless, comparison with available systems at similar actuation parameters places \textit{OmniFish} in a favorable position with respect to energetic performance (Figure~\ref{fig:sota-comp}a).

Several highly optimized robotic fish achieve higher absolute swimming speeds, particularly platforms designed around a single locomotion strategy. However, \textit{OmniFish} prioritizes multimodal capability while maintaining a mechanically simple single-actuator architecture. Despite this design objective, \textit{OmniFish} demonstrated competitive energetic performance relative to several soft robotic fish systems. Notably, the Ostraciiform configuration achieved lower COT than highly optimized platforms such as \textit{Tunabot}~\cite{zhu2019tuna}, while all investigated modes had lower COT than \textit{SoFi}~\cite{katzschmann2018exploration}. These results suggest that the present tendon-driven architecture can be energetically favorable compared with several soft robotic systems relying on pneumatic or fluidic actuation, while preserving a comparatively simple mechanical design.

To the authors’ knowledge, only two previous robotic systems have demonstrated multiple BCF locomotion modes within a single platform~\cite{wang2025soft,wang2026adaptive}. \textit{OmniFish} extends this concept by incorporating five BCF-inspired locomotion modes, including Ostraciiform swimming, within a unified morphology and single-actuator design. Comparison with these multimodal systems further highlights the tradeoff between versatility and performance (Figure~\ref{fig:sota-comp}b). The MDEA-driven platform~\cite{wang2025soft} achieved comparable performance in Anguilliform, Subcarangiform, and Thunniform locomotion, while the present Ostraciiform implementation exceeded its reported performance. In contrast, the tensegrity-based platform~\cite{wang2026adaptive} achieved substantially higher swimming speed and efficiency, indicating the performance potential of more highly optimized multimodal architectures.

Overall, the comparison suggests that \textit{OmniFish} prioritizes locomotion versatility, mechanical simplicity and energetic efficiency rather than maximizing peak swimming speed. Its ability to emulate five BCF-inspired modes within a single morphology and single-actuator architecture provides a foundation for future adaptive robotic fish capable of selecting propulsion strategies according to task requirements and environmental conditions.

\subsection{Limitations and Future Work}

The experimentally achieved swimming speeds remained substantially below the biological envelope defined in section~\ref{sec:hydro}. This difference is unlikely to result from a single source of excessive drag or insufficient thrust, but rather from the combined influence of the experimental configuration, actuator limitations, losses within the compliant transmission, and that the robot does not directly reproduce the BCF modes but rather emulates them. The present system remains tethered and relies on external guidance through a linear rig. Communication, power supply, and directional guidance are externally imposed, introducing additional constraints and disturbances including cable interaction and carriage drag. Likewise, the current actuator represents a significant limitation. The servo constrained operation to a maximum frequency of 2~TB~s$^{-1}$ while simultaneously reducing achievable caudal amplitudes at higher frequencies. These limitations also prevented the assembled robot from accessing the high-thrust operating conditions identified during isolated fin testing as the selected averaged fin achieved its maximum measured thrust at 7~TB~s$^{-1}$. As discussed in section~\ref{sec:disc-mech}, part of the actuator input was also dissipated through deformation of the compliant spine and silicone body, while the discrete rib-to-skin interfaces may have reduced the of body-wave transmission from the spine. Together, these factors likely limited forward speed and resulted in operation at Reynolds numbers below the biological design envelope and Strouhal numbers above those typically associated with efficient cruising. Since $St=\frac{fH}{U}$, the measured range of 1.38–10.24 primarily reflects the low forward velocity relative to the imposed caudal-fin frequency and peak-to-peak amplitude. The robot therefore generated substantial lateral tail motion without a proportional increase in forward translation, indicating operation outside the kinematic regime associated with efficient biological cruising. However, Strouhal number alone does not quantify propulsive efficiency or the fraction of energy lost to lateral flow. The present study was not designed to separate the individual contributions of tether loading, carriage friction, tendon transmission, silicone deformation, or hydrodynamic drag, and these contributions therefore cannot be quantified from the current measurements. The compliant fin–body–tendon system may also exhibit mode-dependent hysteresis through viscoelastic deformation, transmission friction, and phase lag between actuator motion and caudal-fin response. These effects were not measured directly in the present study and therefore cannot be separated from the other mechanical losses.

Future work should combine untethered operation with improved actuator bandwidth and transmission efficiency. Removing the linear guide associated with the external loading in the current test configuration and incorporating onboard power, control, communication, and buoyancy management, while maintaining the robot's soft, reconfigurable architecture,  would provide a more representative assessment of locomotion performance. Alternative actuators or transmission systems could maintain larger caudal-fin amplitudes at higher frequencies, shift optimum operating points, and further improve performance. Brushless DC motors combined with oscillatory transmission mechanisms represent one possible future direction such as \textit{Tunabot Flex}~\cite{white2021tunabot}. Bringing the system closer to the biological Strouhal range will require joint optimization of frequency, caudal amplitude, forward speed, and the stiffness or elastic modulus of the spine, skin, and fin, rather than independently reducing frequency or amplitude. Future characterization should therefore combine Strouhal number and COT with direct measurements of drivetrain efficiency, net thrust, and wake structure. Finally, mode transitions currently require partial disassembly of the posterior body and manual insertion of jamming blocks. The present platform should therefore be regarded as manually reconfigurable between experiments rather than capable of dynamic mode switching during operation. This requirement limits its immediate use for tasks that demand real-time adaptation to changing flow conditions, confined environments, or mission requirements. Translating the current proof-of-concept into an actively switchable stiffness system is therefore an important next step. Potential implementations include pneumatically actuated jamming structures, distributed SMA elements, or longitudinal deployable stiffening mechanisms. Such systems would need to provide sufficient locking stiffness while remaining compact, waterproof, low-power, and compatible with the compliant body architecture. Future designs should also enable stiffness switching without disturbing tendon tension or requiring removal of the external skin.

Future experiments should also evaluate maneuverability, turning radius, turning rate, acceleration, station keeping, disturbance rejection, and swimming in cluttered or confined environments. These tests are essential for determining whether the lower-speed undulatory modes provide functional advantages beyond straight-line propulsion. In addition, flow visualization and force measurements during free swimming could provide deeper insight into how each mode generates thrust and how stiffness distribution affects wake structure, energy transfer, propulsive efficiency and hysteresis.

Overall, \textit{OmniFish} demonstrates by proof-of-concept that morphology averaging combined with mechanically configurable stiffness can provide a compact and versatile strategy for multimodal soft robotic fish. While the current implementation favors localized tail propulsion for straight-line swimming, the ability to emulate the full BCF mode spectrum within a single platform provides a useful foundation for future adaptive underwater robots capable of selecting locomotion strategies according to task and environment.

\section{Conclusion}
This work presented \textit{OmniFish}, a modular, manually reconfigurable soft robotic fish designed to span the full range of BCF locomotion modes within a single robotic platform. The robot combines an averaged fish-inspired body morphology, a soft caudal fin selected through thrust characterization, and a tendon-driven compliant spine with mechanically configurable stiffness. By progressively jamming spinal segments, the robot emulates Anguilliform, Subcarangiform, Carangiform, Thunniform, and Ostraciiform locomotion using a single actuator. Table~\ref{tab:omnifish_specs} show the specifications for \textit{OmniFish}.

\begin{table}[htbp]
\centering
\caption{OmniFish specification sheet.}
\label{tab:omnifish_specs}
\begin{tabular}{ll}
\hline
Attribute & Value \\
\hline
Size & $L$: 45.42~cm, $W$: 5.66~cm $D$: 9.67~cm \\
Robot volume & 1.019~L \\
Robot mass & 1.065~kg (flooded), 0.343~kg (dry) \\
Locomotion modes & Anguilliform, Subcarangiform, Carangiform, Thunniform, Ostraciiform \\
Transmission & Antagonistic tendon drive \\
Tested frequencies & 0.6, 1.0, 1.5, 2.0~TB~s$^{-1}$ \\
Swimming speed (max) & 8.14~cm~s$^{-1}$, Ostraciiform at 1.5~TB~s$^{-1}$ \\
Normalized speed (max) & 0.179~BL~s$^{-1}$ \\
Lowest COT & 2.42, Ostraciiform at 1.5~TB~s$^{-1}$ \\
Best mode & Ostraciiform \\
\hline
\end{tabular}
\end{table}

The design process integrated morphometric analysis, caudal fin thrust characterization, numerical body-drag simulations, robot implementation, kinematic analysis, and swimming performance experiments. The averaged caudal fin produced the highest thrust among the tested fin designs, while the averaged body achieved favorable drag characteristics and reduced sensitivity to off-axis flow. Kinematic analysis confirmed that progressive segment locking shifted deformation posteriorly, emulating the intended transition from undulatory to oscillatory BCF-inspired locomotion, but also revealed distinct differences from biological kinematics such as amplitude envelope and phase progression. Swimming characterization showed a clear dependence of performance on both body stiffness and actuation frequency. Contrary to initial expectations, the Ostraciiform configuration achieved the best straight-line performance, reaching 8.14~cm~s$^{-1}$ with a COT of 2.42 at 1.5~TB~s$^{-1}$. These results suggest that the present architecture mechanically favors localized tail propulsion for straight-line swimming, while more compliant undulatory modes may offer functional advantages in maneuverability and environmental interaction that remain to be explored..

Overall, \textit{OmniFish} demonstrates that morphology averaging and stiffness reconfiguration can provide a compact and mechanically simple strategy for developing multimodal soft robotic fish. Although the current implementation relies on manual stiffness reconfiguration and tethered straight-line experiments, it establishes a proof-of-concept foundation for future adaptive underwater robots capable of dynamically selecting locomotion strategies according to task requirements and environmental conditions. Future work will focus on active stiffness switching, untethered operation, improved actuation bandwidth, and evaluation of maneuverability, turning, and swimming in cluttered or realistic aquatic environments.
%
%


\funding{The authors gratefully acknowledge the financial support of the Thomas B. Thrige Fond and the Brødrene Hartmanns Fond under their equipment grant program.}

\roles{S.P.M.B. conceived the conceptual idea. M.J. and S.P.M.B. did the study and developed the overall research direction. M.J. designed and fabricated the robotic platform, performed the morphometric analysis, caudal-fin characterization, CFD simulations, kinematic experiments, swimming experiments, and data analysis. S.P.M.B. supervised the project, contributed to the conceptualization of the bioinspired design strategy, guided the experimental design and interpretation of results, and provided resources and project administration. M.J. prepared the initial manuscript draft. M.J. and S.P.M.B. discussed the results, revised the manuscript, and approved the final version.}


\data{All 3D design files, control code, data, and data analysis used in the manuscript is available through \url{https://github.com/SDUSoftRobotics/OmniFish}}

\suppdata{See the attachment}

\bibliographystyle{unsrtnat}
\bibliography{sources}

@article{rajeshkumar2018studies,
  title={Studies on seasonal pollution of heavy metals in water, sediment, fish and oyster from the Meiliang Bay of Taihu Lake in China},
  author={Rajeshkumar, Sivakumar and Liu, Yang and Zhang, Xiangyang and Ravikumar, Boopalan and Bai, Ge and Li, Xiaoyu},
  journal={Chemosphere},
  volume={191},
  pages={626--638},
  year={2018},
  publisher={Elsevier}
}

@article{wynn2014autonomous,
  title={Autonomous Underwater Vehicles (AUVs): Their past, present and future contributions to the advancement of marine geoscience},
  author={Wynn, Russell B and Huvenne, Veerle AI and Le Bas, Timothy P and Murton, Bramley J and Connelly, Douglas P and Bett, Brian J and Ruhl, Henry A and Morris, Kirsty J and Peakall, Jeffrey and Parsons, Daniel R and others},
  journal={Marine geology},
  volume={352},
  pages={451--468},
  year={2014},
  publisher={Elsevier}
}

@incollection{Fileman_deMora_Vance_2020, 
title={Physical Effects of Ships on the Environment}, 
booktitle={Environmental Impact of Ships}, 
publisher={Cambridge University Press}, 
author={Fileman, Timothy and de Mora, Stephen and Vance, Thomas}, 
editor={de Mora, Stephen and Fileman, Timothy and Vance, Thomas}, 
year={2020}, 
pages={216--224}
}

@article{cai2025measuring,
  title={Measuring and Minimizing Disturbance of Marine Animals to Underwater Vehicles},
  author={Cai, Levi and J{\'e}z{\'e}quel, Youenn and Mooney, T Aran and Girdhar, Yogesh},
  journal={arXiv preprint arXiv:2506.11335},
  year={2025}
}

@Book{springer2016Handbook,
  editor    = {Siciliano, Bruno AND Khatib, Oussama},
  publisher = {Springer International Publishing},
  title     = {Springer Handbook of Robotics, 2\textsuperscript{nd} Edition},
  year      = {2016},
}

@article{fish2020advantages,
  title={Advantages of aquatic animals as models for bio-inspired drones over present AUV technology},
  author={Fish, Frank E},
  journal={Bioinspiration \& biomimetics},
  volume={15},
  number={2},
  pages={025001},
  year={2020},
  publisher={IOP Publishing}
}

@article{webb1984form,
  title={Form and function in fish swimming},
  author={Webb, Paul W},
  journal={Scientific American},
  volume={251},
  number={1},
  pages={72--83},
  year={1984},
  publisher={JSTOR}
}

@article{sfakiotakis1999review,
  title={Review of fish swimming modes for aquatic locomotion},
  author={Sfakiotakis, Michael and Lane, David M and Davies, J Bruce C},
  journal={IEEE Journal of oceanic engineering},
  volume={24},
  number={2},
  pages={237--252},
  year={1999},
  publisher={IEEE}
}

@article{triantafyllou2000hydrodynamics,
  title={Hydrodynamics of fishlike swimming},
  author={Triantafyllou, Michael S and Triantafyllou, George S and Yue, Dick KP},
  journal={Annual review of fluid mechanics},
  volume={32},
  number={1},
  pages={33--53},
  year={2000},
  publisher={Annual Reviews 4139 El Camino Way, PO Box 10139, Palo Alto, CA 94303-0139, USA}
}

@incollection{LAUDER2005425,
title = {Hydrodynamics of Undulatory Propulsion},
series = {Fish Physiology},
publisher = {Academic Press},
volume = {23},
pages = {425-468},
year = {2005},
booktitle = {Fish Biomechanics},
issn = {1546-5098},
author = {George V. Lauder and Eric D. Tytell},
}

@article{crespi2005amphibot,
  title={AmphiBot I: an amphibious snake-like robot},
  author={Crespi, Alessandro and Badertscher, Andr{\'e} and Guignard, Andr{\'e} and Ijspeert, Auke Jan},
  journal={Robotics and Autonomous Systems},
  volume={50},
  number={4},
  pages={163--175},
  year={2005},
  publisher={Elsevier}
}

@article{hultmark2007flowfield,
  title={Flowfield measurements in the wake of a robotic lamprey},
  author={Hultmark, Marcus and Leftwich, Megan and Smits, Alexander J},
  journal={Experiments in fluids},
  volume={43},
  number={5},
  pages={683--690},
  year={2007},
  publisher={Springer}
}

@inproceedings{milana2020eelworm,
  title={EELWORM: a bioinspired multimodal amphibious soft robot.},
  author={Milana, Edoardo and Van Raemdonck, Bert and Cornelis, Kevin and Dehaerne, Enrique and De Clerck, Jef and De Groof, Yarno and De Vil, Toon and Gorissen, Benjamin and Reynaerts, Dominiek},
  booktitle={RoboSoft},
  pages={766--771},
  year={2020}
}

@article{nguyen2022anguilliform,
  title={Anguilliform swimming performance of an eel-inspired soft robot},
  author={Nguyen, Dinh Quang and Ho, Van Anh},
  journal={Soft Robotics},
  volume={9},
  number={3},
  pages={425--439},
  year={2022},
  publisher={SAGE Publications Sage CA: Los Angeles, CA}
}

@article{christianson2018translucent,
  title={Translucent soft robots driven by frameless fluid electrode dielectric elastomer actuators},
  author={Christianson, Caleb and Goldberg, Nathaniel N and Deheyn, Dimitri D and Cai, Shengqiang and Tolley, Michael T},
  journal={Science Robotics},
  volume={3},
  number={17},
  pages={eaat1893},
  year={2018},
  publisher={American Association for the Advancement of Science}
}

@article{katzschmann2018exploration,
  title={Exploration of underwater life with an acoustically controlled soft robotic fish},
  author={Katzschmann, Robert K and DelPreto, Joseph and MacCurdy, Robert and Rus, Daniela},
  journal={Science Robotics},
  volume={3},
  number={16},
  pages={eaar3449},
  year={2018},
  publisher={American Association for the Advancement of Science}
}

@article{muralidharan2021development,
  title={Development of subcarangiform bionic robotic fish propelled by shape memory alloy actuators},
  author={Muralidharan, Mohanakrishnan and Palani, Iyamperumal Anand},
  journal={Defence Science Journal},
  volume={71},
  number={1},
  pages={94--101},
  year={2021}
}

@article{shintake2018soft,
  title={Soft biomimetic fish robot made of dielectric elastomer actuators},
  author={Shintake, Jun and Cacucciolo, Vito and Shea, Herbert and Floreano, Dario},
  journal={Soft robotics},
  volume={5},
  number={4},
  pages={466--474},
  year={2018},
  publisher={SAGE Publications Sage CA: Los Angeles, CA}
}

@article{wang2023design,
  title={Design, fabrication, and realisation of a robotic fish actuated by dielectric elastomer with a passive fin},
  author={Wang, Zekai and Lou, Junqiang and Xiao, Xingdong and Li, Guoping and Deng, Yimin},
  journal={IET Cyber-Systems and Robotics},
  volume={5},
  number={2},
  pages={e12090},
  year={2023},
  publisher={Wiley Online Library}
}

@article{youssef2022design,
  title={Design and control of soft biomimetic pangasius fish robot using fin ray effect and reinforcement learning},
  author={Youssef, Samuel M and Soliman, MennaAllah and Saleh, Mahmood A and Elsayed, Ahmed H and Radwan, Ahmed G},
  journal={Scientific Reports},
  volume={12},
  number={1},
  pages={21861},
  year={2022},
  publisher={Nature Publishing Group UK London}
}

@article{zhu2019tuna,
  title={Tuna robotics: A high-frequency experimental platform exploring the performance space of swimming fishes},
  author={Zhu, Joseph and White, Carl and Wainwright, Dylan K and Di Santo, Valentina and Lauder, George V and Bart-Smith, Hilary},
  journal={Science Robotics},
  volume={4},
  number={34},
  pages={eaax4615},
  year={2019},
  publisher={American Association for the Advancement of Science}
}

@article{white2021tunabot,
  title={Tunabot Flex: A tuna-inspired robot with body flexibility improves high-performance swimming},
  author={White, Carl H and Lauder, George V and Bart-Smith, Hilary},
  journal={Bioinspiration \& Biomimetics},
  volume={16},
  number={2},
  pages={026019},
  year={2021},
  publisher={IOP Publishing}
}

@article{zhong2021tunable,
  title={Tunable stiffness enables fast and efficient swimming in fish-like robots},
  author={Zhong, Qiang and Zhu, Joseph and Fish, Frank E and Kerr, Steven John and Downs, AM and Bart-Smith, Hilary and Quinn, DB},
  journal={Science Robotics},
  volume={6},
  number={57},
  pages={eabe4088},
  year={2021},
  publisher={American Association for the Advancement of Science}
}

@article{van2022openfish,
  title={OpenFish: Biomimetic design of a soft robotic fish for high speed locomotion},
  author={Van Den Berg, Sander C and Scharff, Rob BN and Rus{\'a}k, Zolt{\'a}n and Wu, Jun},
  journal={HardwareX},
  volume={12},
  pages={e00320},
  year={2022},
  publisher={Elsevier}
}

@misc{xiong2024carbonfishbistableunderactuated,
      title={CarbonFish -- A Bistable Underactuated Compliant Fish Robot capable of High Frequency Undulation}, 
      author={Zechen Xiong and Zihan Guo and Mark Liu and Jialong Ning and Hod Lipson},
      year={2024},
      eprint={2311.03223},
      note={arXiv:2311.03223},
      archivePrefix={arXiv},
      primaryClass={cs.RO},
}

@article{kodati2008microautonomous,
  title={Microautonomous robotic ostraciiform (MARCO): Hydrodynamics, design, and fabrication},
  author={Kodati, Parasar and Hinkle, Jonathan and Winn, Aaron and Deng, Xinyan},
  journal={IEEE Transactions on Robotics},
  volume={24},
  number={1},
  pages={105--117},
  year={2008},
  publisher={IEEE}
}

@article{wang2025soft,
  title={Soft robotic fish actuated by bionic muscle with embedded sensing for self-adaptive multiple modes swimming},
  author={Wang, Ruiqian and Zhang, Chuang and Tan, Wenjun and Zhang, Yiwei and Yang, Lianchao and Chen, Wenyuan and Wang, Feifei and Tian, Jiandong and Liu, Lianqing},
  journal={IEEE Transactions on Robotics},
  volume={41},
  pages={1329--1345},
  year={2025},
  publisher={IEEE}
}

@article{wang2026adaptive,
  title={Adaptive multimodal swimming gaits in a reconfigurable modular soft robotic fish},
  author={Wang, Bo and Li, Lei and Xu, Mengfan and Hu, Nannan and Gao, Wenzhuo and Zhang, Jie and Yin, Bo and Xin, Zhanhua and Yu, Junzhi},
  journal={Science Advances},
  volume={12},
  number={1},
  pages={eaea1299},
  year={2026},
  publisher={American Association for the Advancement of Science}
}

@article{palstra2008cost,
  title={Cost of transport and optimal swimming speed in farmed and wild European silver eels (Anguilla anguilla)},
  author={Palstra, Arjan and van Ginneken, Vincent and van den Thillart, Guido},
  journal={Comparative Biochemistry and Physiology Part A: Molecular \& Integrative Physiology},
  volume={151},
  number={1},
  pages={37--44},
  year={2008},
  publisher={Elsevier}
}

@article{webb1971swimming,
  title={The swimming energetics of trout: I. Thrust and power output at cruising speeds},
  author={Webb, Paul W},
  journal={Journal of experimental biology},
  volume={55},
  number={2},
  pages={489--520},
  year={1971},
  publisher={The Company of Biologists Ltd}
}

@article{sambilay1990interrelationships,
  title={Interrelationships between swimming speed, caudal fin aspect ratio and body length of fishes},
  author={Sambilay Jr, VICTOR C},
  journal={Fishbyte},
  volume={8},
  number={3},
  pages={16--20},
  year={1990},
  publisher={The WorldFish Center}
}

@article{hove2001boxfishes,
  title={Boxfishes (Teleostei: Ostraciidae) as a model system for fishes swimming with many fins: kinematics},
  author={Hove, JR and O’Bryan, LM and Gordon, Malcolm S and Webb, Paul W and Weihs, Daniel},
  journal={Journal of Experimental Biology},
  volume={204},
  number={8},
  pages={1459--1471},
  year={2001},
  publisher={The Company of Biologists Ltd}
}

@article{arakeri2009fluid,
  title={Fluid mechanics of fish swimming},
  author={Arakeri, Jaywant H},
  journal={Resonance},
  volume={14},
  number={1},
  pages={32},
  year={2009}
}

@article{di2021convergence,
  title={Convergence of undulatory swimming kinematics across a diversity of fishes},
  author={Di Santo, Valentina and Goerig, Elsa and Wainwright, Dylan K and Akanyeti, Otar and Liao, James C and Castro-Santos, Theodore and Lauder, George V},
  journal={Proceedings of the National Academy of Sciences},
  volume={118},
  number={49},
  pages={e2113206118},
  year={2021},
  publisher={National Academy of Sciences}
}

@inproceedings{walter1996approximate,
  title={Approximate arc length parameterization},
  author={Walter, Marcelo and Fournier, Alain},
  booktitle={Proceedings of the 9th Brazilian symposium on computer graphics and image processing},
  pages={143--150},
  year={1996},
  organization={SBC/UFMG Caxambu}
}

@article{hu2021micro,
  title={Micro thrust measurement experiment and pressure field evolution of bionic robotic fish with harmonic actuation of macro fiber composites},
  author={Hu, Dongyi and Lou, Junqiang and Chen, Tehuan and Yang, Yiling and Xu, Chao and Chen, Hairong and Cui, Yuguo},
  journal={Mechanical systems and signal processing},
  volume={153},
  pages={107538},
  year={2021},
  publisher={Elsevier}
}

@book{versteeg1995finitevolume,
  author = {Versteeg, Henk K. and Malalasekera, Weeratunge},
  isbn = {0582218845},
  publisher = {Pearson Prentice Hall},
  title = {An introduction to computational fluid dynamics},
  year = 1995
}

@article{menter1994two,
  title={Two-equation eddy-viscosity turbulence models for engineering applications},
  author={Menter, Florian R},
  journal={AIAA journal},
  volume={32},
  number={8},
  pages={1598--1605},
  year={1994}
}

@book{Anderson2011Fundamentals,
  author = {Anderson, John D.},
  edition = {5th},
  publisher = {McGraw-Hill},
  title = {{Fundamentals of aerodynamics}},
  year = 2011
}

@inproceedings{ta2024single,
  title={Single Actuator Undulation Soft-bodied Robots Using A Precompressed Variable Thickness Flexible Beam},
  author={Ta, Tung D},
  booktitle={2024 IEEE/RSJ International Conference on Intelligent Robots and Systems (IROS)},
  pages={255--261},
  year={2024},
  organization={IEEE}
}

@article{akanyeti2013kinematic,
  title={A kinematic model of K{\'a}rm{\'a}n gaiting in rainbow trout},
  author={Akanyeti, Otar and Liao, James C},
  journal={Journal of Experimental Biology},
  volume={216},
  number={24},
  pages={4666--4677},
  year={2013},
  publisher={Company of Biologists}
}

\end{document}